\documentclass[journal]{IEEEtran}
\usepackage[pdftex]{graphics} 
\usepackage{epsfig} 
\usepackage{times} 
\usepackage{amsmath} 
\usepackage{amssymb}  
\usepackage{bm}
\usepackage{subfigure}
\usepackage[T1]{fontenc}
\usepackage[linesnumbered,ruled,vlined]{algorithm2e}
\usepackage[noend]{algpseudocode}
\usepackage{booktabs}
\usepackage{array}
\usepackage{color}
\usepackage[colorlinks,linkcolor=red,anchorcolor=blue,citecolor=green]{hyperref}
\usepackage{multirow}

\usepackage{float}
\usepackage{xcolor}
\usepackage{url}
\usepackage{cite}
\usepackage[export]{adjustbox}
\usepackage{balance}

\begin{document}
\title{Real-time 3D Single Object Tracking\\ with Transformer}

\author{Jiayao Shan$^\dagger$, Sifan Zhou$^\dagger$, Yubo Cui$^\dagger$, Zheng Fang*,~\IEEEmembership{Member,~IEEE,}
	\thanks{$^\dagger$Authors with equal contribution.}
	\thanks{The authors are with the Faculty of Robot Science and Engineering, Northeastern University, Shenyang, China; Corresponding author: Zheng Fang, e-mail: fangzheng@mail.neu.edu.cn}%
	\thanks{$^{2}$Jiayao Shan is also with Science and Technology on Near-Surface Detection Laboratory, Wuxi, China}%
	\thanks{This paper has supplementary downloadable material available at http://ieeexplore.ieee.org, provided by the authors.}
	}	

\markboth{Journal of \LaTeX\ Class Files,~Vol.~14, No.~8, August~2015}%
{Shell \MakeLowercase{\textit{et al.}}: Bare Demo of IEEEtran.cls for IEEE Journals}

\maketitle

\begin{abstract}
LiDAR-based 3D single object tracking is a challenging issue in robotics and autonomous driving. Currently, existing approaches usually suffer from the problem that objects at long distance often have very sparse or partially-occluded point clouds, which makes the features extracted by the model ambiguous. Ambiguous features will make it hard to locate the target object and finally lead to bad tracking results. To solve this problem, we utilize the powerful Transformer architecture and propose a \textbf{Point-Track-Transformer (PTT)} module for point cloud-based 3D single object tracking task. Specifically, PTT module generates fine-tuned attention features by computing attention weights, which guides the tracker focusing on the important features of the target and improves the tracking ability in complex scenarios. To evaluate our PTT module, we embed PTT into the dominant method and construct a novel 3D SOT tracker named PTT-Net. In PTT-Net, we embed PTT into the voting stage and proposal generation stage, respectively. PTT module in the voting stage could model the interactions among point patches, which learns context-dependent features. Meanwhile, PTT module in the proposal generation stage could capture the contextual information between object and background. We evaluate our PTT-Net on KITTI and NuScenes datasets. Experimental results demonstrate the effectiveness of PTT module and the superiority of PTT-Net, which surpasses the baseline by a noticeable margin, $\sim$10\% in the Car category. Meanwhile, our method also has a significant performance improvement in sparse scenarios. In general, the combination of transformer and tracking pipeline enables our PTT-Net to achieve state-of-the-art performance on both two datasets. Additionally, PTT-Net could run in real-time at 40FPS on NVIDIA 1080Ti GPU. Our code is open-sourced for the research community at \url{https://github.com/shanjiayao/PTT}.
\end{abstract}


\IEEEpeerreviewmaketitle

\section{Introduction}
\label{sec:introduction}
\IEEEPARstart{S}{ingle} object tracking (SOT) using LiDAR points has a wide range of applications in robotics and autonomous driving \cite{comport2004robust,kartobject,machida2012human}. For example, the autonomous pedestrian following robot should robustly track its master and localize him/her accurately for efficient following control in the crowd. Another example is autonomous landing of unmanned aerial vehicles, where the drone needs to track the target and know the accurate distance and pose of the target for safe landing. However, most existing 3D SOT methods are usually using visual or RGB-D cameras \cite{Bibi3D,context,RGBDtracker}, which may fail in visually degraded or illumination changing environments due to that they mainly depend on the dense images for target tracking.

In addition to visual or RGB-D sensors, 3D LiDAR sensors are also widely used in object tracking tasks \cite{FaF,Complexer-YOLO,pointtracknet} because they are less sensitive to illumination changes and could directly capture geometric and distance information more accurately. However, LiDAR-based 3D SOT has its own challenges. \textit{First}, point data is sparse and disordered \cite{PointNet}, which requires the network to be permutation-invariant to handle points well. \textit{Second}, point cloud is spatially discrete, which is naturally different from dense image. \textit{Third}, 3D object tracking needs to estimate higher space dimension information (e.g., $x,y,z,w,h,l,ry$) than 2D visual tracking, which brings more computational complexity. All these problems bring great challenges to realize a robust and real-time LiDAR-based tracking method.

\begin{figure}[t]
	\centering
    \setlength{\abovecaptionskip}{-6pt}    
	\includegraphics[width=\linewidth]{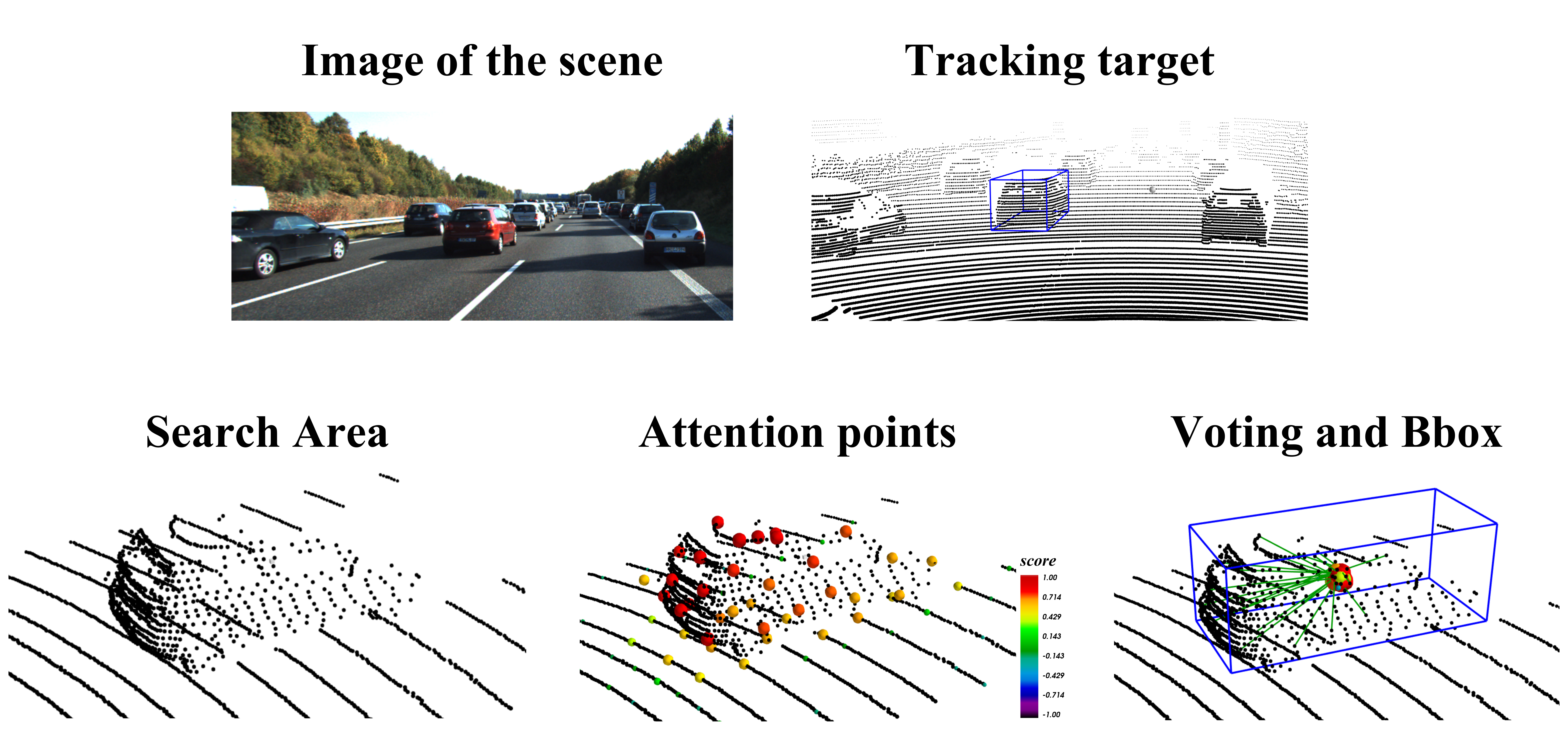}
	\caption{\textbf{Exemplified illustration to show the attention weights and voting result}. Given raw point cloud, we specify search area and extract robust key-points first. Then, the transformer and attention mechanism focus on the key-points which carry rich and robust information. Finally, the voting module will generate votes according to the key-points and the predict bounding box.}    
	\label{fig:abstract}
\end{figure}

Different from the existing LiDAR-based Multi-object Tracking (MOT) methods~\cite{ab3dmot,PointTrackNetAE,Shenoi2020JRMOTAR,Baser2019FANTrack3M,Luo2018}, LiDAR-based 3D SOT methods need to model the similarity function between target template and search area to localize the target object. Although they both need to compute the similarity, MOT methods compute the object-level similarity to association the detection results and tracklets, while SOT methods compute the intra-object-level similarity to localize the target object. Therefore, compared to 3D MOT, 3D SOT has its own challenges. SC3D \cite{SC3D} is the pioneer LiDAR-based 3D Siamese tracker which is based on the shape completion network. However, the method only uses an encoder consisting of 3 layers of 1D-convolutions to process the input point cloud, which makes it difficult to extract the robust point cloud feature representation. Besides, SC3D could not run in real-time and be trained end-to-end. Qi et al. \cite{P2B} also proposed a point-to-box (P2B) network to estimate the target bounding box from the raw point cloud. However, their approach usually fails to track in sparse point cloud scenarios. Meanwhile, P2B gives no preference to non-accidental coincidences \cite{SPOT} which have more contribution to locating the target center. Recently, Fang et al. \cite{3DSiamRPN} combined Siamese network and LiDAR-based RPN network \cite{PointRCNN} to tackle 3D object tracking. Nevertheless, they directly use the classification scores to sort regression results, ignoring the inconsistency between the localization and classification. It is worth noting that points located in different geometric positions often have different importance in representing targets. However, these aforementioned methods do not weigh point cloud features based on this characteristic. Besides, the point cloud features extracted from the template and the search area contain less potential object information and more background noise due to the sparsity and occlusion of point clouds. Therefore, how to pay attention to the spatial clues is the key to improve the performance of the 3D object tracker.

Recently, Transformer has shown amazing performance in feature encoding due to its powerful self-attention module\cite{vaswani2017,wu2019pay,Devlin2019}. Transformer usually consists of three main modules, including input (word) embedding, position encoding and attention module. Compared with the convolution network, content-adaptive property and unique position module of transformer make it more suitable for processing 3D point clouds. In addition, the 3D SOT task only focuses on local areas, which makes the transformer suitable for this task, although the transformer is sensitive to time and space cost.

In this paper, we propose a Point-Track-Transformer (PTT) module for 3D single object tracking to learn features more effectively by leveraging the superiority of the transformer models on set-structured point clouds. The core idea is to focus on the important features of the target object by utilizing the self-attention and position encoding mechanism to weigh the point cloud features. PTT module contains three blocks for feature embedding, position encoding, and self-attention feature computation, respectively. Feature embedding aims to place features closer in the embedding space if they have similar semantic information. Position encoding is used to encode the coordinates of point cloud into high dimension distinguishable features. Self-attention generates refined attention features by computing attention weights. Furthermore, to evaluate the effectiveness of our PTT module, we embed the PTT module into the dominant P2B \cite{P2B} to construct a novel 3D SOT tracker termed PTT-Net. In PTT-Net, we add PTT into the voting stage and proposal generation stage, respectively. PTT embedded in the voting stage could model interactions among point patches located in different geometric positions, which learns context-dependent features and helps the network focus on more representative features of objects. Meanwhile, PTT embedded in the proposal generation stage could capture the contextual information between object and background, and help the network to effectively suppress background noise. These modifications can efficiently improve the performance of 3D object tracker. The experimental results of our PTT-Net on KITTI tracking dataset~\cite{kitti} demonstrate the superiority of our method ($\sim$10\%'s improvement compared to the baseline). We further evaluate our PTT-Net on NuScenes dataset~\cite{nuscenes}, the results show that our method could achieve new state-of-the-art performance. Additionally, PTT-Net could run in real-time at 40FPS on a single NVIDIA 1080Ti GPU. 

Overall, our main contributions are as follows:

$\bullet$ \textbf{PTT module}: we propose a Point-Track-Transformer (PTT) module for 3D single object tracking using only raw point clouds, which could weigh point cloud features to focus on deeper-level object clues during tracking.

$\bullet$ \textbf{PTT-Net}: we construct a 3D single object tracking network embedded with PTT modules which can be trained end-to-end. To the best of our knowledge, this is the first work to apply transformer to 3D object tracking task using point clouds.

$\bullet$ \textbf{Open-source}: extensive experiments on KITTI and NuScenes datasets show that our method outperforms the state-of-the-art methods with remarkable margins at the speed of 40 FPS. Besides, we open source of our method to the research community.

Apart from that, as an extended work of our conference paper \cite{shan2021ptt}, we add more detailed descriptions on the network architecture and dataset. Besides, we also carry out more qualitative experiments and visualizations of our PTT-Net to analyze the effectiveness of transformer in 3D single object tracking task. For the extended experiments on the more challenging NuScenes dataset, our method still achieves state-of-the-art performance. The results indicate that our method could be adapted to more complex scenes, further confirming the effectiveness of our method.

The rest of this paper is organized as follows. In Sec. II,
we discuss the related work. Section III describes the proposed PTT module and PTT-Net. We validate the performance of our method on KITTI and NuScenes datasets in Sec. IV and we conclude in section V.

\section{RELATED WORK}
\label{sec:related-works}

This section will briefly discuss the related works in 2D siamese trackers, 3D single object tracking, transformer, and self-attention mechanism.

\subsection{2D Siamese Tracking}
Early 2D visual trackers mainly focused on correlation filtering \cite{Bolme2010,KCF,ECO,Galoogahi2015}. However, these methods are based on the tracking template matching mechanism, so they cannot cope with the rapid deformation of the tracking target. Recently, the realization of 2D object tracking tasks based on Siamese networks has become the mainstream with the rapid development of deep learning~\cite{SiamFC,SiameseRPN,SiamRPN++,Dong2018TripletLI,SiamMask,siamrcnn,EnhancedLoss,Han2021LearningTF,Dong2020CLNetAC,Dong2021DynamicalHO}. Luca et al. \cite{SiamFC} proposed SiamFC which was the first pioneer work of Siamese trackers. The visual tracking task was handled as a similarity problem, and the cross-correlation module was introduced into the network structure. Subsequently, a large number variants of SiamFC \cite{SiamFC} were proposed. Li et al. \cite{SiameseRPN} introduced the Region Proposal Network (RPN) into the Siamese network. SiamRPN could regress more accurate 2D bounding box than SiamFC. Besides, Li et al. \cite{SiamRPN++} explored the relationship between the number of network layers and tracker performance, optimized the network depth of the tracker to improve the tracking accuracy. Furthermore, Wang et al. \cite{SiamMask} integrated the task framework of image segmentation and image tracking, and used the mask to improve the accuracy of the tracker. Paul et al. \cite{siamrcnn} proposed a two-stage network for visual tracking, which used the re-detection of the first frame template and the previous frame template to modify the tracking target. Their method surpassed all previous methods on six short-term tracking benchmarks and four long-term tracking benchmarks, and achieved amazing results. \cite{EnhancedLoss} introduced an informative enhanced loss, which can enable the network to capture information from an overall perspective. Han et al.~\cite{Han2021LearningTF} proposed an  asymmetric convolution module, which could capture the semantic correlation information well. To address the problem of decisive samples missing during offline training, Dong et al.~\cite{Dong2020CLNetAC} proposed a compact latent network to make the model could quickly adapt to new scenes. And Dong et al.~\cite{Dong2021DynamicalHO} introduced a novel hyper-parameter optimization method by using deep reinforcement learning. Chen et al.~\cite{transformertrack} proposed a tracking framework by fusing the template and search features with transformer. In summary, the 2D visual tracking method has made great progress in the past decade and has been applied to many practical scenarios. However, limited by the sensor, the 2D visual trackers are still very sensitive to illumination changes. In addition, most of the 2D visual tracking methods only obtain the pixel coordinates of the tracking target, but sometimes it is necessary to know the accurate three-dimensional pose of the tracking target.

\subsection{3D SOT Using Point Cloud}
Giancola et al. \cite{SC3D} proposed the first pioneer LiDAR-based 3D single object tracker which utilized the Kalman Filter to generate massive target proposals. They exploited shape completion to learn the shape information of target, but their method has a poor generalization ability and could not run in real-time. Zarzar et al. \cite{Zarzar2019} leveraged 2D Siamese network which works on Bird-Eye-View (BEV) representation to generate 3D proposals. However, this method may lose fine-grained geometry details which are important for tracking tiny objects. Cui et al. \cite{Cui2019} also adopted a 3D Siamese tracker only using point cloud, but they could not estimate the orientation and size information of the target. Fang et al. \cite{3DSiamRPN} combined 3D Siamese network and 3D RPN network to track targets, while the performance is limited by the RPN network. Besides, Zou et al. \cite{FSiamese} integrated 2D image and 3D point cloud information for 3D object tracking. However, this method relies more on 2D tracker and uses the ground truth to track objects, which is unreasonable for realistic application. Qi et al. \cite{P2B} proposed P2B which used deep hough voting to obtain the potential centers (votes) and estimated the target center based on those votes. However, they ignore the fact of points in different positions have different contributions in tracking. Furthermore, its random sampling mechanism loses the location distribution information of the raw point cloud. To deal with these shortcomings, we propose a PTT module to capture the feature correlations among the neighbor point around the target object by weighing different point features. Moreover, we use farthest point sampling instead of random sampling to obtain more raw point cloud information.

\subsection{Transformer and Self-attention}
Recently, transformer has revolutionized natural language processing and image analysis \cite{vaswani2017,hu2019local,ramachandran2019stand,Carion2020,vqa}. Hu et al. \cite{hu2019local} and Ramachandran et al. \cite{ramachandran2019stand} applied scalar dot product self-attention to local pixel neighbors. Zhao et al. \cite{zhao2020san} applied vector self-attention operations to image tasks. These works combined or replaced CNNs with self-attention layers and confirmed the transformer's great potential in visual tasks. 

Inspired by those works, Zhao et al. \cite{Jia22020} used a Point Transformer layer by applying vector self-attention operations, which had a great performance improvement in point cloud classification and segmentation tasks. Because self-attention operator, which is the core of transformer networks, is intrinsically a set operator: positional information is provided as attributes of elements that are processed as a set \cite{vaswani2017,zhao2020san}. Therefore, transformer is suitable for point cloud processing due to its positional attributes. Besides, Nico et al. \cite{Engel2020} proposed SortNet as a part of Point Transformer and achieved competitive performance on point cloud classification and part segmentation tasks. Meanwhile, Guo et al. \cite{Guo2020} also introduced Point Cloud Transformer (PCT), which performed well on shape classification, part segmentation, and normal estimation tasks. Recently, Pan et al. \cite{Pan2020} proposed a PointFormer as the drop-in replacement backbone for 3D object detection and gained state-of-the-art performance. Obviously, transformer has unique advantages for point cloud feature learning.

In addition, transformer and attention mechanism have recently been widely used in 2D tracking tasks \cite{TrackFormer,TransTrack,Chu2021,transformertrack,TransformerMT,Shen2020VisualOT}. The tracker using the transformer or attention also shows superior performance with the help of the transformer's powerful attention mechanism for features. Therefore, we apply the transformer to the 3D point cloud tracking task to improve the performance of the tracker.

\section{METHODOLOGY}
\label{sec:method}

In this section, we first analyze the challenges and bottlenecks of current LiDAR-based single object trackers and discuss feasible solutions. Then, we revisit the transformer and present our PTT module for LiDAR-based object tracking. Finally, we introduce our PTT-Net in detail. 

\subsection{Baseline}
P2B~\cite{P2B} is the dominant 3D SOT method using point clouds. In this work, we use P2B as the baseline. The main idea of P2B is to localize the target center in 3D search area and execute the proposal generation and verification jointly. Hence we can divide P2B into two parts. The first part is feature enhancement. The input template and search point clouds are extracted through a shared weight backbone network~\cite{PointNet++}, then the corresponding similarity could be obtained by calculating the point-wise cosine similarity in an implicitly embedded space. The second part is the region proposal network which generates the proposals by deep hough voting mechanism~\cite{VoteNet} from the semantic features. Besides, P2B also utilizes the proposal clustering network to leverage the ensemble power and obtain accurate target proposals.

However, P2B tends to suffer from the defects that it gives no preference to non-accidental coincidences \cite{SPOT} which have more contribution to locate the target center. Therefore, we would like to explore the differences among the augmented features by using the transformer architecture.

\subsection{Challenges}
\label{sec:challenges}
In Sec.~\ref{sec:introduction}, we had pointed out several challenges for processing point cloud data. For 3D single object tracking task, there are also several challenges as follows.

\subsubsection{error accumulation and propagation}
In tasks involving point clouds and deep learning, tracking is naturally different from detection and segmentation due to its spatial-temporal continuity. Since the tracking target is a sequence, which makes two frames similar in spatial and temporal terms. Thus, the dominant tracking algorithms utilize the spatial and temporal prior information to initialize the search area. In spite of its effectiveness in reducing computational complexity, bad tracking results will lead to large tracking error in challenging scenarios. This is because the consecutive tracking predictions will accumulate over the historical error and propagate it to the next frame. And when error accumulates enough, the tracker will fails.

\begin{figure*}[ht]
	\centering
	\setlength{\abovecaptionskip}{-20pt}
	\includegraphics[width=\linewidth]{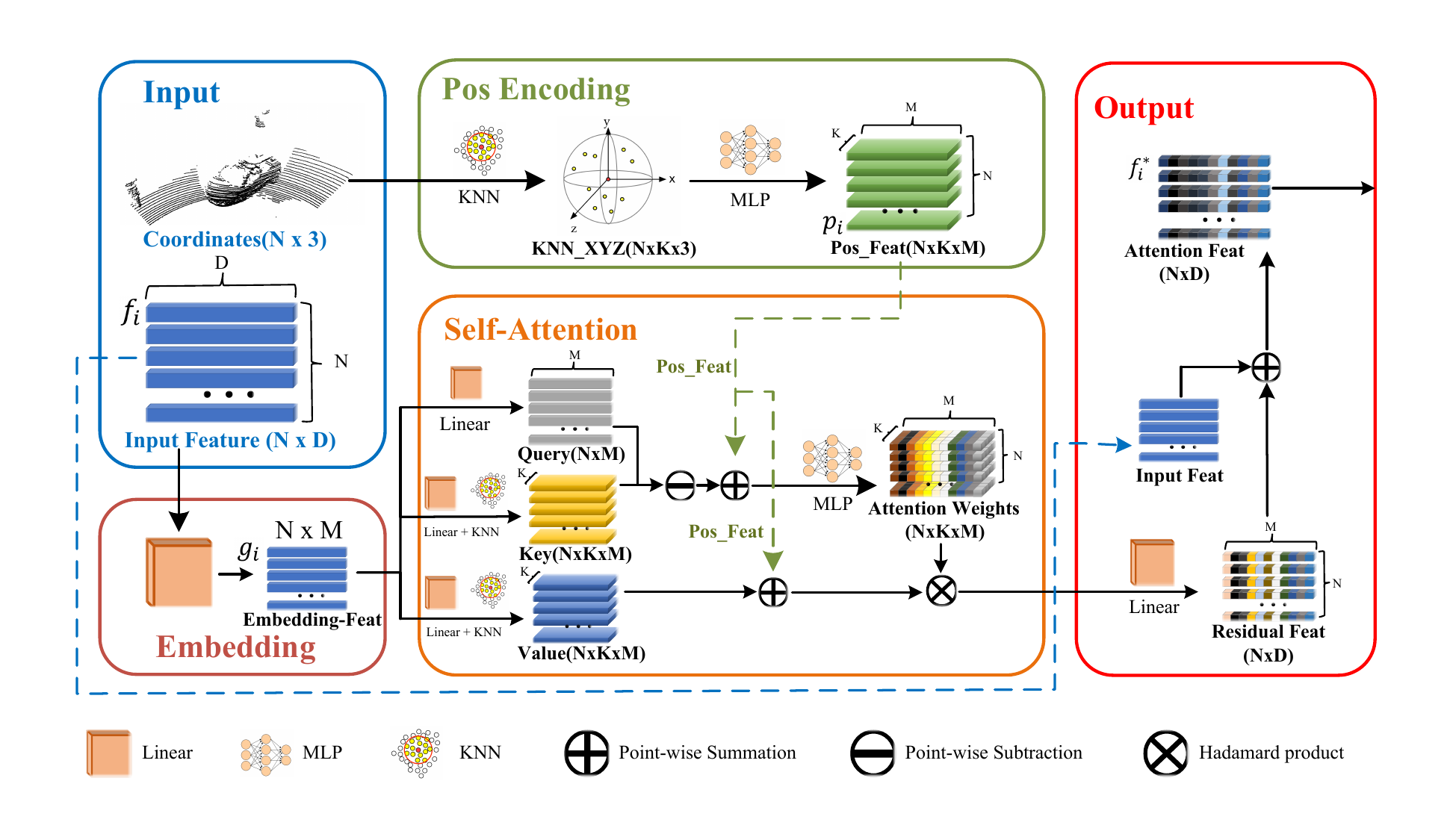}
	\caption{\textbf{PTT module architecture.} It consists of three blocks: feature embedding, position encoding, and self-attention. The whole input are the coordinates and their corresponding features. Feature embedding module maps input features into embedding space. In position encoding module, the k-nearest neighbor algorithm is used to obtain local position information, then the encoded position features will be learned by an MLP layer. The self-attention module learns refined attention features for input features based on local context. The output features of PTT module are the sum of input and residual features.}
	\label{fig:transformer}
\end{figure*}

\subsubsection{sparsity sensitive}
The Siamese network from VOT is generally adopted in existing algorithms for point cloud object tracking. However, the image data is usually dense, while point cloud is naturally sparse. This causes gaps in the effect of applying Siamese network in image and point cloud. Sparse point cloud data makes it difficult for the backbone network to extract robust point cloud features. Hence, the existing LiDAR-based tracking methods are sensitive to the sparsity of point cloud data.

\subsubsection{feature ambiguity}
The sparsity of point clouds limits networks on modeling interactions among point patches located in different geometric positions and capturing contextual information, which makes the features extracted by the models ambiguous. 
The points at locations where the object surface have a low-dimensional structure, such as a plane, contribute ambiguous features\cite{SPOT}. The feature ambiguity makes the tracker hard to classify the fore-background points and regress the box center, and finally leads to bad tracking results.

These challenges motivate us to propose new algorithm to deal with those problems. To this end, we propose our PTT (Point-Track-Transformer) module and PTT-Net. PTT module can handle the sparsity of points with the help of the attention mechanism in Transformer, where the contribution of each point is automatically learned in the network training. Besides, the ambiguity of features will be suppressed through self-attention mechanism because of the role of attention is to refine features and guide the network to focus on more representative features of the tracking target. Finally, our PTT-Net embedded with PTT module could capture sparse dependencies even from a few points, thus reducing the accumulation of errors.

\subsection{Revisiting Transformer}
Transformer \cite{vaswani2017} is firstly introduced to aggregate information from the entire input sequence for machine translation. It can handle sequential tasks well due to its attention mechanism. The core architecture of transformer can be divided into three parts: input feature embedding, position encoding, and self-attention. Self-attention is the core module, which mainly focuses on the differences of input features and generates refined attention features based on global or local context. Given the input feature ${G=\left\{g_i\right\}^{N}_{i=1}}$ after feature embedding, the general formula of self-attention is:
\begin{equation}
    \begin{aligned}
        &Q,K,V = \alpha(G), \beta(G), \gamma(G)\\
        &\mathcal{A} = \rho(\sigma(Q, K) + P) \odot (V)
    \end{aligned}
\label{eq:general-trans}
\end{equation}
where $\alpha$, $\beta$ and $\gamma$ are point-wise feature transformations (e.g. linear layers or MLPs). $Q$, $K$, and $V$ are the $query$, $key$ and $value$ matrices, respectively. $\sigma$ is the relation function between $Q$ and $K$. $P$ is the position encoding feature. $\rho$ is a normalization function (e.g. \textit{softmax}). $\odot$ means Hadamard product, which is used to obtain the output features from the attention weights and $V$. $\mathcal{A}$ is the attention feature produced by the self-attention module. For the relation function $\sigma$, the regular form in machine translation \cite{vaswani2017} is:
\begin{equation}
    \begin{aligned}
        \sigma(Q, K) = QK^T
    \end{aligned}
\label{eq:standard-sigma}
\end{equation}

\subsection{PTT Module}
To further integrate the self-attention mechanism into the point cloud tracking task, we modify the transformer module proposed in \cite{Jia22020} to capture point cloud features better. The point transformer layer in \cite{Jia22020} is proposed to process the raw point cloud for classification and segmentation tasks. However, we utilize the point transformer to benefit the tracking task and enable the tracker to capture spatial relations and object geometry shape information. All these modifications construct the PTT module, which is used to refine the features from raw sparse point clouds and eliminate the ambiguity among features. The architecture of PTT module is shown in Fig.~\ref{fig:transformer}.

For 3D SOT task, given an input of M points with XYZ coordinates, a backbone network is used to extract the point cloud and learn deep features. It outputs a subset of the input containing N interest points (seeds) ${S=\left\{s_i\right\}^{N}_{i=1}}$. Here,  ${s_i=(c_i,f_i)}$ is composed of a vector ${c_i}$ of 3D coordinate and a D-dimensional descriptor ${f_i}$ of the local object geometry. Our goal of using transformer is to perform an attention weighting operation on the feature space of $f_i$, and output refined features ${f_i}^*$ with the same dimension.

PTT module processes features by utilizing shape and geometry information. Given a point set ${S=\left\{s_i\right\}^{N}_{i=1} }$, $s_i=(c_i,f_i)$, $c_i \in \mathbb{R}^{3}$ and $f_i \in \mathbb{R}^{D}$. ${c_i}$ and ${f_i}$ represent 3D coordinates and descriptor of point $s_i$. Feature embedding module maps input features into embedding space $\mathbb{R}^{M}$: $f_i \rightarrow g_i$, $g_i \in \mathbb{R}^{M}$. Position encoding module extracts higher-level M-dimensional features $p_i$ from input coordinates $c_i$: $c_i \rightarrow p_i$, $p_i \in \mathbb{R}^{K \times M}$. Finally, the self-attention module calculates attention weights and attention features $f_i^*$,$f_i^* \in \mathbb{R}^{D}$ by taking embedding features and position features as inputs. To avoid the vanishing gradient problem in the training stage, we also adopt the residual architecture proposed in \cite{resnet}, and take the sum of the attention features and input features as output features.

\subsubsection{Feature Embedding}
The original feature embedding module in natural language processing is to map each word in the input sequence to a high-dimensional vector. In this work, we use the linear layer to complete the feature embedding operation, and map the input point cloud feature dimension from $D$ To $M$: $\mathbb{R}^{D} \rightarrow \mathbb{R}^{M}$, which can place the feature closer in the embedding space when the semantics are more similar and make the network have a stronger fitting ability.

\begin{figure*}[t]
	\centering
	\setlength{\abovecaptionskip}{-16pt}
	\includegraphics[width=\linewidth]{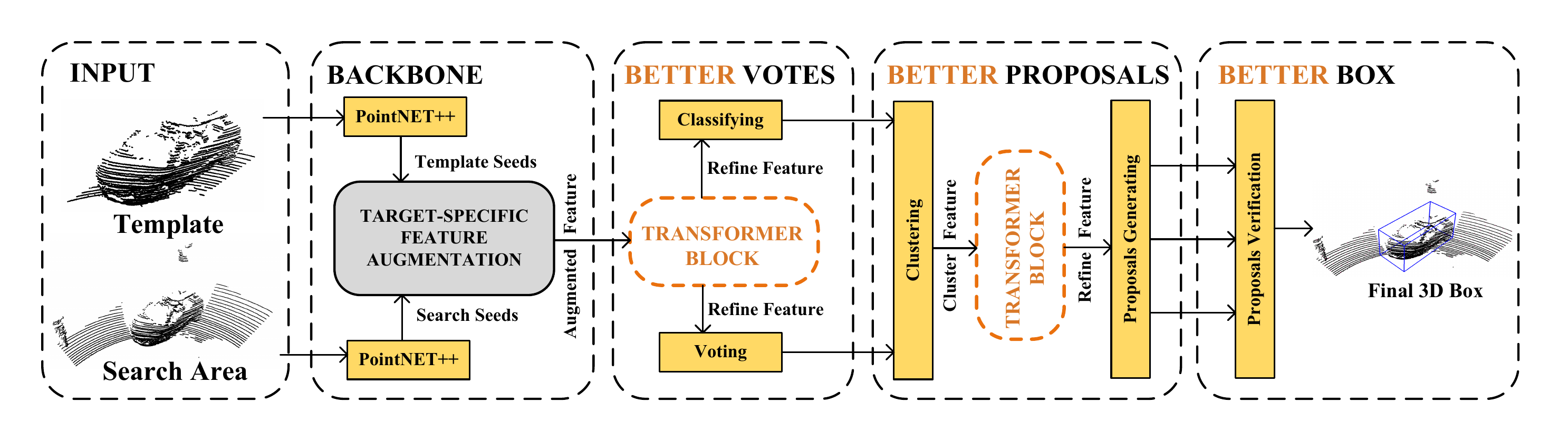}
	\caption{\textbf{The pipeline of PTT-Net.} In order to verify the effect of our PTT module, we embed two PTT modules into seeds voting and proposals generation stage of the deep hough voting framework.}
	\label{fig:main-pipeline}
\end{figure*}

\subsubsection{Position Encoding}
Position encoding module plays a crucial role in transformer, which allows operators to adapt to the local structure of the input data \cite{vaswani2017}. 3D point coordinates are valuable features indicating the local structures. Therefore, we utilize the coordinates directly as the input of the position encoding module. 
Compared to the techniques used in natural language processing, we use a simple and yet efficient approach by mapping the coordinates of each point to the feature dimension and the resulting position encoding is added to the attention matrix. We adopt the relative coordinates to make the network better capture the spatial correlation between points and local geometric shape information. Since the feature $f_i$ extracted by \cite{PointNet++} can provide the local context information, we obtain the position encoding features ${P=\left\{p_i\right\}^{N}_{i=1}}$ with function $\eta$. For input point set $S$ including N points, the position encoding feature for each point is:
\begin{equation}\label{eq:pos-encode}
    {p_i}= \eta(c_i - c_j)
\end{equation}
where $c_i$ is the coordinate of the i-th point in $S$. $c_j$ is the j-th coordinate in local neighborhood region of $c_i$. $\eta$ is an MLP with two linear layers and one ReLU non-linearity. Here, we use $K$NN to capture the local context and set $k$ = 16 by inheriting the experimental results in \cite{Jia22020}.

\subsubsection{Self-Attention} 
As Fig.~\ref{fig:transformer} shows, self-attention module computes three vectors for each point: $Q$, $K$, $V$ through $\alpha$, $\beta$, $\gamma$, where $\alpha$, $\beta$, $\gamma$ are all shared linear layers. It is worth noting that $K$ and $V$ are aggregated from the features of the $k$ neighborhood points, which aims to encode more local context information. Here, $Q \in \mathbb{R}^{M}$, $K \in \mathbb{R}^{k \times M}$, and $V \in \mathbb{R}^{k \times M}$. For relation function $\sigma$, we use $\sigma(Q, K)$ to obtain point-wise attention weights, the detail implementation of $\sigma(Q, K)$ will be introduced in next part. And an MLP layer $\gamma$ is used to introduce additional trainable transformations and match the output dimension. Then, we add the position encoding features $P$ to both the attention vector $\sigma$ and the transformed features $K$. Finally, the residual features recorded as $\mathcal{A}$ are defined as the weighted sum of the attention weights with all $V$ vectors. The formula is as follows:
\begin{equation}\label{eq:attention}
\mathcal{A} = \rho\big(\gamma(\sigma(Q, K) + {P})\big) \odot \big({V} + {P}\big)
\end{equation}
where $\rho$ is a normalization function (softmax) and $\gamma$ is a non-linear mapping function (MLP) that includes two linear layers and one ReLU non-linearity. $\sigma$ is the relation function between $Q$ and $K$. $\mathcal{A}$ is output attention features.

\subsubsection{Relations Functions}
\label{sec:relation-func}
The relation function $\sigma(Q, K)$ is the core of self-attention. Different ways of obtaining the relation between Query vector and Key vector could construct different types of attention modules. As mentioned in \cite{Jia22020}, self-attention operators can be classified into two types: scalar attention \cite{vaswani2017} and vector attention \cite{zhao2020san}. In scalar attention, the relation function $\sigma(Q, K)$ can be expressed in the form of Eq.~\ref{eq:standard-sigma}, which computes the scalar product between Query vector and Key vector. The vector attention obtains vector attention weights by using channel-wise subtraction operation. Besides, it has been confirmed in \cite{Jia22020} that vector attention is a natural fit for point cloud than scalar attention since it supports adaptive modulation of individual feature channels, not just whole feature vectors. Therefore, we use the vector attention structure. The relation function $\sigma(Q, K)$ is a subtraction operation. The formula is as follows:

\begin{equation}\label{eq:basic-attention}
\begin{aligned}
    &\sigma(Q, K) = Q - K
\end{aligned}
\end{equation}

\subsection{PTT-Net}
\label{sec:ptt-net}
This section details our PTT-Net which is a more accurate and robust target tracking based on the existing open-source method P2B \cite{P2B}. In the following, we first explore the effect of different sampling methods on point cloud-based tracking tasks. Then, the position where the PTT module is embedded  and the loss function of the network training are described.

\begin{figure}[t]
	\centering
	\setlength{\abovecaptionskip}{-5pt}
	\includegraphics[width=\linewidth]{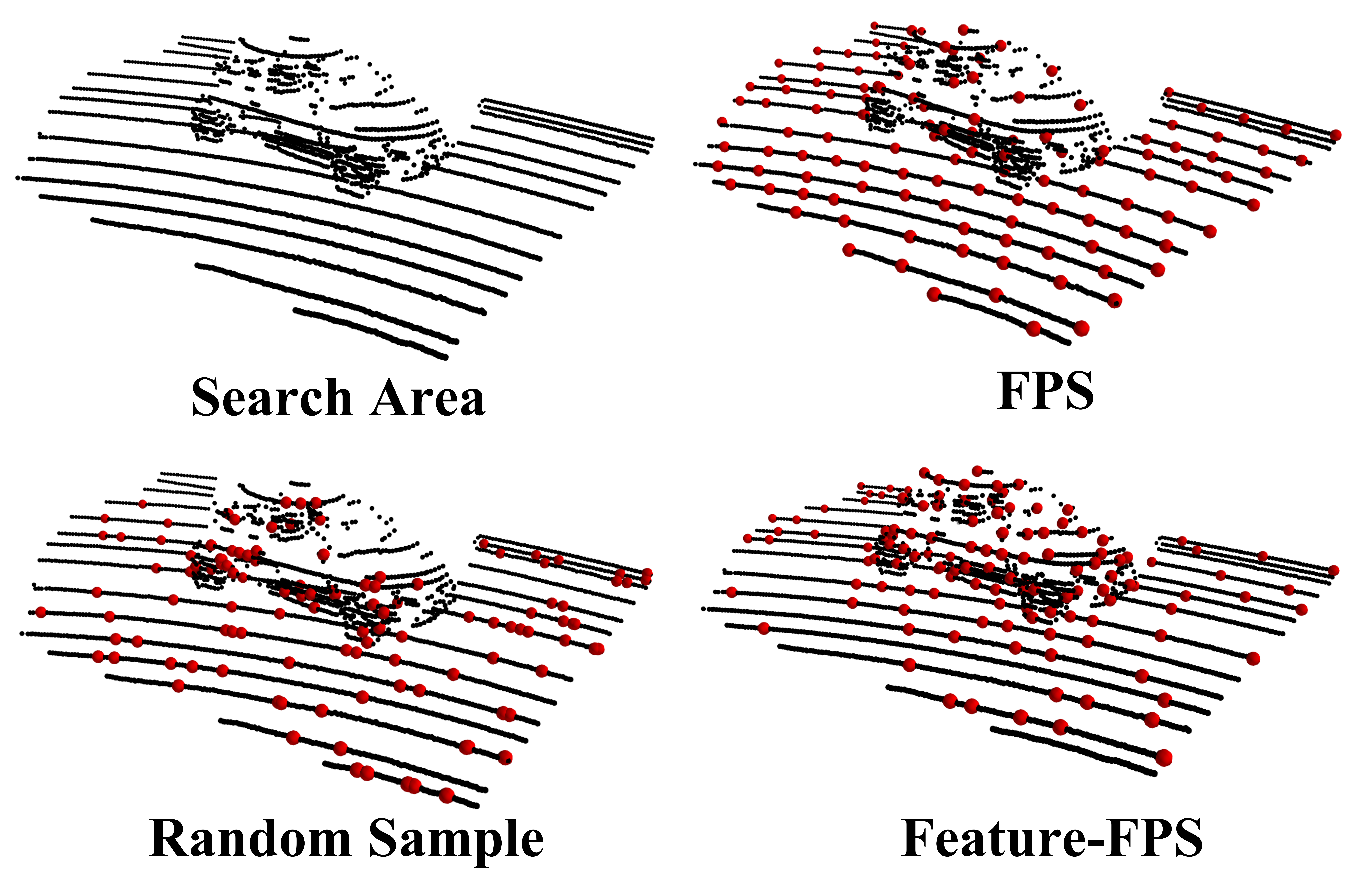}
	\caption{\textbf{Visualization of different sampling methods}. The input points and sampled points are labeled as black and red respectively. As shown in the figure, the result of RS depends on the distribution density. And FPS can better retain the geometric information. However, the Feat-FPS mainly focuses on the foreground points, which will cause the unbalanced distribution between foreground and background points.}
	\label{fig:sample_method}
\end{figure}

\subsubsection{Sampling Strategies}\label{sec:sample}
Here, we first discuss the impact of different sampling strategies on tracking task. The purpose of sampling is to extract key points and ensure that the number of points in the template or search point cloud is aligned with the input dimension of network. However, different sampling methods will lead to different degrees of target information loss. At the same time, we find that in the existing tracking pipeline, the more foreground points left in the search area after sampling, the more accurate the regression results of the network. In contrast, due to the unbalanced distribution of foreground and background points, the classification accuracy of the network will decrease. Therefore, a suitable sampling strategy can not only achieve input alignment, but also improve the tracking performance of the tracking network. 

Common 3D point cloud sampling methods include \textit{Random Sampling (RS)}, \textit{Farthest Point Sampling (FPS)}, and \textit{Farthest Point Sampling in Feature dimension (Feat-FPS)}. RS can achieve high sampling efficiency, but the sampling result depends on the distribution density of point cloud. FPS can better retain the geometric information of the original point cloud, and is a more balanced sampling method. In addition, Feat-FPS proposed in 3D-SSD \cite{3DSSD}, which can sample in feature space, has better sampling results for targets of different semantic categories. The visualization of the three different sampling methods is shown in Fig.~\ref{fig:sample_method}. We find that FPS has more uniform sampling results, while Feat-FPS pays more attention to points on the target. P2B uses RS, although it has a higher computational efficiency, it leads to the loss of some key information, which limits their tracking accuracy. 

In summary, for the sampling strategy of tracking task, we adopted FPS which can retain the geometric information of the original point cloud and make the foreground and background points balanced. A demonstration of its good tracking performance will be shown in Sec.~\ref{sec:ablation-study}.

\subsubsection{Embedding Position of PTT}
\begin{figure}[t!]
	\centering
	\setlength\abovecaptionskip{-10pt}
	\includegraphics[width=\linewidth]{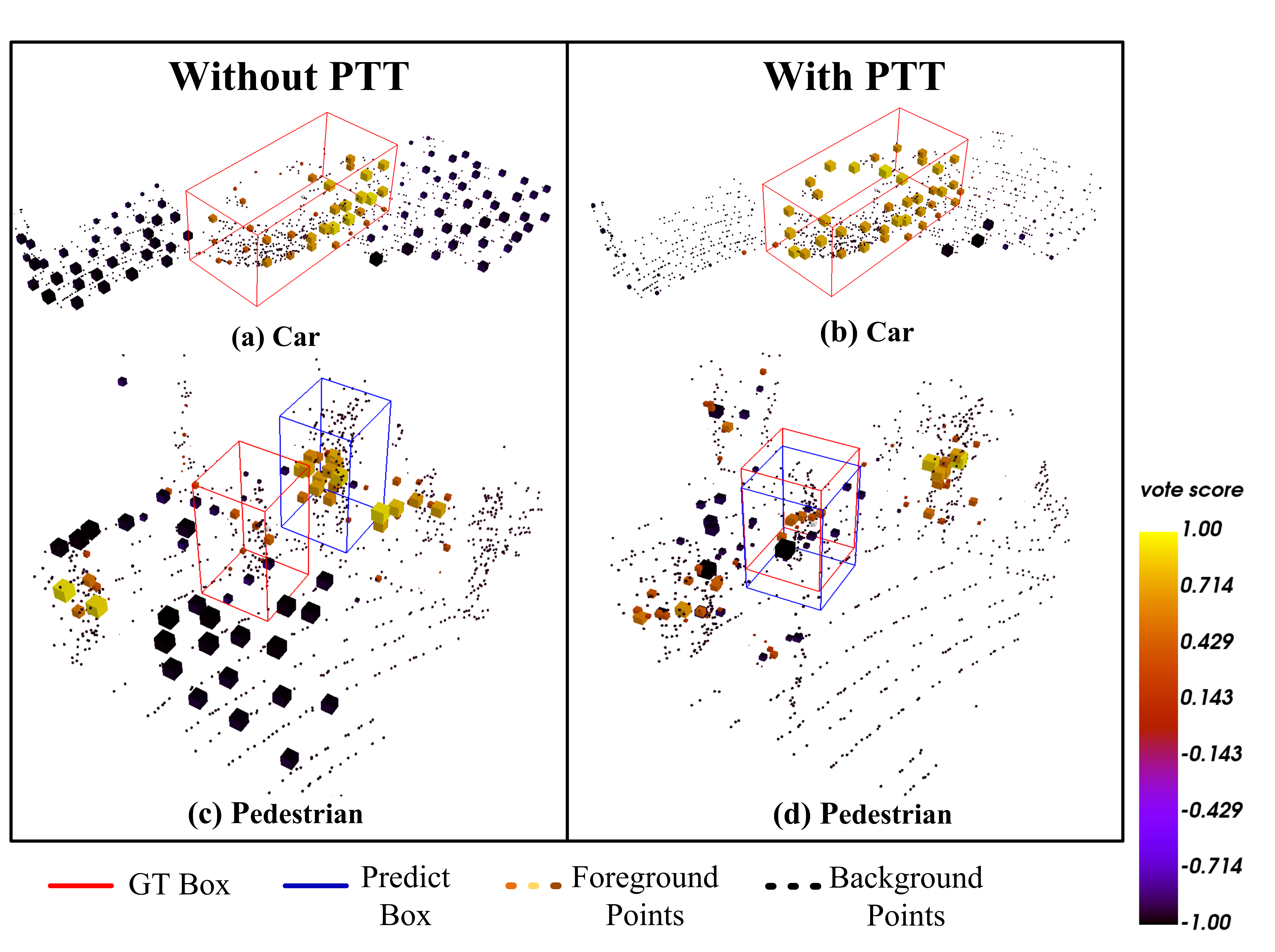}
    \caption{\textbf{Visualization of classification(a-b) and tracking(c-d) results with or without PTT module}. The point will be paid more attention if it has a higher score. Compared (a) with (b), PTT module pays more attention to the foreground points. Compared (c) with (d), PTT module could still track targets robustly in crowded scenes(with multiple pedestrians).}
	\label{fig:transformer_show}
\end{figure}

The ability of transformer to learn self-attention weights inspires us to try it on 3D SOT task. We formulate the problem of focusing on the differences of features as self-attention weighting. In order to verify the effect of our method, we embed our PTT module into the previous open-source \textit{state-of-the-art} (SOTA) LiDAR-based 3D SOT work P2B \cite{P2B}. More specifically, the PTT modules are inserted in seeds voting stage and proposals generation stage of P2B respectively.
In the seeds voting stage, P2B generates votes by utilizing the augmented features, which are obtained from the backbone network (in Fig.~\ref{fig:main-pipeline}). We notice that \cite{P2B} ignores the differences among different point cloud features in the search area, and gives no preference to the points in different locations when generating votes. However, it is important to focus on the points which contain more geometric information. Therefore, we apply PTT module to weigh the augmented features and model interactions among point patches to learn the context-dependent feature (in Fig.~\ref{fig:transformer_show}(a)(b)). 

In proposals generation stage, P2B generates proposals based on local context features. However, their method ignores the global semantic features of the targets, so that they could not distinguish similar objects (e.g., two pedestrians, in Fig.~\ref{fig:transformer_show}(c)(d)). Therefore, we use the PTT module to further weigh the target-wise context features obtained by the aggregation network in P2B for tracking deeper-level target clues to capture the contextual information between object and background.

As shown in Fig.~\ref{fig:main-pipeline}, we embed our PTT module in the open-source SOTA method P2B \cite{P2B} to build PTT-Net. We add PTT module to the seeds voting and proposal generation stages, and weigh the augmented features and cluster features respectively. Experiments show that our PTT-Net outperforms the SOTA method with remarkable margins.

\subsubsection{Loss Function}
The PTT module is trained with the other sub-networks in \cite{P2B}. Therefore, we follow \cite{P2B} to design our loss function. The overall loss consists of two parts as follows:
\begin{equation}
\begin{aligned}
&{L}_{all} = {L}_{cv} + \lambda_{1}{L}_{cb}+\lambda_{2}{L}_{rv}+\lambda_{3}{L}_{rb}
\end{aligned}
\label{equ:loss}
\end{equation}
where $\lambda_{1}$, $\lambda_{2}$, $\lambda_{3}$ represent the weighting coefficient of each loss. Classification loss includes voting classification loss ${L}_{cv}$ and proposal box classification loss ${L}_{cb}$. The regression loss includes the voting loss ${L}_{rv}$ and the proposal box regression loss ${L}_{rb}$.

\begin{figure}[t]
\begin{center}
	\centering
	\setlength\abovecaptionskip{-10pt}
	\includegraphics[width=1.0\linewidth]{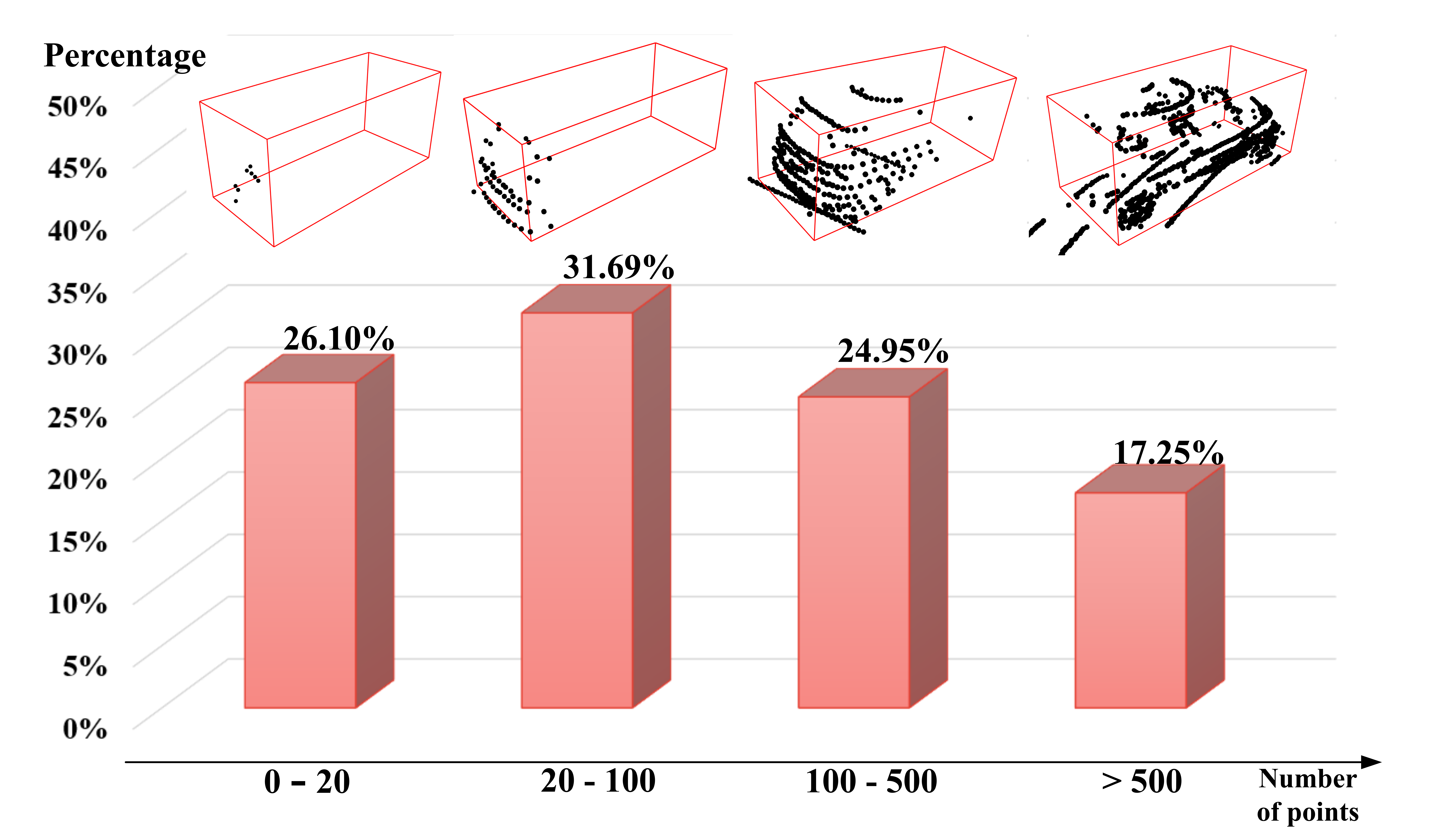}
	\caption{\textbf{The percentages and visualizations of the different number intervals of point cloud in the car category.} The number of frames containing less than 20 points accounts for 26.10\% of the total. The number of frames containing 20-100 points accounts for 31.69\% of the total. In addition, the number of 100-500 and more than 500 frames are accounting for 24.95\% and 17.25\% of the total, respectively. For four different intervals, we visualize four point clouds with 11, 52, 293, and 883 points, respectively.}
	\label{fig:dataset}
\end{center}
\end{figure}

\begin{table*}[t]
\renewcommand\arraystretch{1.3}
    \begin{center}
    \caption{Performance comparison on \textbf{KITTI} for the \textbf{car} category. Red and blue mean the performance score is ranked first and second respectively.}
    \label{tab:evaluation_car}
    \setlength{\tabcolsep}{27pt}{
	\begin{tabular}{c|c||c|c|c}
		\toprule[.05cm]
		Module  & Modality & 3D Success & 3D Precision & FPS \\\hline
		AVOD-Tracking\cite{AVOD}  & RGB+LiDAR & 63.1 & 69.7 & -\\
		F-Siamese\cite{FSiamese}  & RGB+LIDAR & 37.1 & 50.6 & -\\\hline
		SC3D\cite{SC3D}  & LiDAR only & 41.3 & 57.9 & 1.8\\
		ETP2D-3D\cite{Zarzar2019} & LiDAR only & 36.3 & 51.0 & -\\
		P2B\cite{P2B} & LiDAR only & 56.2 & 72.8 & {\color{red} 45.5}\\
		3D-SiamRPN\cite{3DSiamRPN} & LiDAR only & {\color{blue} 58.2} & {\color{blue} 76.2} & 20.8\\
		PTT-Net(Ours)  & LiDAR only & {\color{red} 67.8} & {\color{red} 81.8} & {\color{blue} 40.0}\\
		\toprule[.05cm]
    \end{tabular}}
    \end{center}
\end{table*}

\section{EXPERIMENTS}
\label{sec:experiment}
We used KITTI tracking dataset \cite{kitti} and NuScenes \cite{nuscenes} dataset as the benchmark. Similar to \cite{SC3D,P2B,3DSiamRPN,FSiamese}, we mainly focused on rigid car tracking and performed ablation studies on KITTI. We also conducted extended experiments with other three target categories (Pedestrian, Van, Cyclist) to comprehensively evaluate the performance of our method for non-rigid objects (Pedestrian) tracking on KITTI. Besides, we also follow BAT \cite{bat} to evaluate our PTT-Net on NuScenes. The experiments show that PTT-Net outperforms previous SOTA methods with remarkable margins and can run at 40 fps. \footnote{Our experiment video is available at \url{https://youtu.be/z5Vkm8r9Wus}.}

\subsection{Experimental Protocols}
\subsubsection{Datasets}
We used the training set of KITTI and NuScenes dataset to train and evaluate our method. For KITTI dataset, there are more than 20,000 manually labeled 3D objects using Velodyne HDL-64E 3D lidar (10HZ). Following \cite{SC3D,P2B,3DSiamRPN,FSiamese}, we split the dataset as follows: 0-16 for training, 17-18 for validation and 19-20 for testing. 
Specially, we first extract each frame label from every scene label of KITTI tracking benchmark. Then, we further extract each ID label from every frame label, and finally concatenate the labels of the same ID to obtain each tracklet label from its first frame to its final frame. By this way, we convert the KITTI MOT label to SOT label. Furthermore, for each tracklet, only the first frame includes the 3D ground truth bounding box (bbox) in the testing phase. Therefore, we initialize our tracker with the first frame 3D ground truth bbox during tracking, and then track the object in the sequence frames by our tracker. In order to better illustrate that the point cloud in KITTI is challenging for tracking task, we choose the Car category and count the number of foreground points in each frame, then calculate the percentage referred to all frames from 00-20 sequence in Fig.~\ref{fig:dataset}. The total number of frames is 27292. The number of frames that are less than 20 points is 7123, accounting for 26.10\% of the total. The number of frames between 20-100 is 8650, accounting for 31.69\% of the total. In addition, the number of 100-500 and more than 500 frames are 6810 and 4709, respectively, accounting for 24.95\% and 17.25\% of the total. This fully shows that about half of the frames in KITTI are sparse scenes. Hence the KITTI dataset is challenging for tracking task. This will be the bottleneck of limiting the tracking accuracy as mentioned in Sec.~\ref{sec:challenges}. Furthermore, we exemplify the visualization results of point clouds at various number intervals in Fig.~\ref{fig:dataset}. 

For NuScenes dataset, there are 1000 driving scenes and 23 object categories, which make it is more challenging because of more complex scenes. We follow the same settings with the BAT \cite{bat} to obtain a fair comparison, and directly refer to the results reported in BAT \cite{bat} for comparison.

\subsubsection{Evaluation Metric}
Following previous work \cite{SC3D,P2B,3DSiamRPN,FSiamese}, we report Success and Precision metrics defined by One Pass Evaluation (OPE) \cite{OPE}, which represent the overlap and error Area Under the Curve (AUC) respectively.

\begin{table*}[t]
\renewcommand\arraystretch{1.3}
	\begin{center}
	    \caption{Extensive comparisons with different categories on \textbf{KITTI}(left) and \textbf{NuScenes}(right) dataset. Red and blue mean the performance score is ranked first and second respectively. And frame number indicates the instance number of each category.}
    	\label{tab:evaluation_KITTI}
		\setlength{\tabcolsep}{8pt}{
		\begin{tabular}{l|c|ccccc|ccccc}
		\toprule[.05cm]
		{ }&Dataset & \multicolumn{5}{c|}{{KITTI}}  & \multicolumn{5}{c}{{NuScenes}}\\ 
		{ }&Category &Car&Pedestrian&Van &Cyclist&Mean & Car & Truck & Trailer & Bus  & Mean \\
		{ }&Frame Number &6424 &6088 &1248 &308 &14068&64159 &13587 &3352&2953 &84051\\\hline
		\multirow{6}{*}{3D Success}
    		&SC3D\cite{SC3D} &41.3 &18.2 &40.4 &{\color{blue}41.5} &31.2 &22.31 &30.67 &35.28 &29.35 &24.43\\
    		&P2B\cite{P2B} &56.2 &28.7 &40.8 &32.1 &42.4 &38.81 &42.95 &48.96 &32.95 &39.68\\
    		&F-Siamese\cite{FSiamese} &37.1 &16.2 &- &{\color{red}47.0} &- &- &- &- &- &-\\
    		&3D-SiamRPN\cite{3DSiamRPN} &58.2 &35.2 &{\color{blue}45.6} &36.1 &46.6 &- &- &- &- &-\\
    		&BAT\cite{bat} &{\color{blue} 65.4} &{\color{red} 45.7} &{\color{red} 52.4} &33.7 &{\color{blue} 55.0} &{\color{blue} 40.73} &{\color{blue} 45.34} &{\color{blue} 52.59} &{\color{blue} 35.44} & {\color{blue} 41.76}\\
    		&PTT-Net(Ours) &{\color{red} 67.8} &{\color{blue} 44.9} &43.6 &37.2 &{\color{red} 55.1} &{\color{red} 41.22} &{\color{red} 50.23} &{\color{red} 61.66} &{\color{red} 43.86} &{\color{red} 43.58}\\
		\hline
		\multirow{5}{*}{3D Precision}
    		&SC3D\cite{SC3D} &57.9 &37.8 &47.0 &{\color{blue}70.4} &48.5 &21.93 &27.73 &28.12 &24.08 &23.19 \\
    		&P2B\cite{P2B}    &72.8 &49.6 &48.4 &44.7 &60.0 &43.18 &41.59 &40.05 &27.41 &42.24\\
    		&F-Siamese\cite{FSiamese} &50.6 &32.2 &- &{\color{red}77.2} &- &- &- &- &- &-\\
    		&3D-SiamRPN\cite{3DSiamRPN}   &76.2 &56.2 &{\color{blue}52.8} &49.0 &64.9 &- &- &- &- &-\\
    		&BAT\cite{bat} &{\color{blue}78.9} &{\color{red}74.5} &{\color{red}67.0} &45.4 &{\color{red}75.2} &{\color{blue}43.29} &{\color{blue}42.58} &{\color{blue}44.89} &{\color{blue}28.01} &{\color{blue}42.70}\\
    		&PTT-Net(Ours) &{\color{red} 81.8} &{\color{blue} 72.0} &52.5 &47.3 &{\color{blue} 74.2} &{\color{red} 45.26} &{\color{red} 48.56} &{\color{red} 56.05} &{\color{red} 39.96} &{\color{red} 46.04}\\
		\toprule[.05cm]
		\end{tabular}
		}
	\end{center}
\end{table*}

\subsubsection{Implementation Details}
We use FPS in our implementation instead of RS used in original P2B \cite{P2B} for point cloud sampling. In the training stage, we use the Adam optimizer and set the initial learning rate to 0.001 and decrease by 5 times after 12 epochs. The batch size is 48 and the training epoch is 60. Besides, we extend the offset from (x, y, $\theta$) to (x, y, z, $\theta$) when generating more template samples during data augmentation in \cite{P2B}. In the testing stage, we also add $Z$ axis offset to generate the predicted box. Other parameters are consistent with settings of \cite{P2B}. Meanwhile, we also follow the tracking setting of P2B~\cite{P2B}. Specially, we initialize the template point cloud with the point cloud of the first frame ground truth and update the template point cloud by fusing the point cloud of first frame ground truth with previous result. The search point cloud is updated based on the point cloud of the previous result, which can better meet the requirement of real scenes.

\subsection{Quantitative Experiment}
To better evaluate our method, we designed two quantitative experiments. In the first experiment, we quantitatively evaluated our method for 3D car tracking. In the second experiment, we further compared PTT-Net with the previous methods among different categories on both KITTI and NuScenes datasets.

\subsubsection{Comparisons on car category}
We compared the performance of our PTT-Net with the existing methods on the KITTI dataset and reported the results for 3D car tracking in Tab.~\ref{tab:evaluation_car}. To meet the requirement of real scenarios, we generate the search area centered on the previous tracking result. The results show that our PTT-Net has achieved SOTA performance in all evaluation metrics with remarkable margins. Compared with the baseline algorithm P2B \cite{P2B}, our performance has been greatly improved by $\sim$11\% in 3D Success. Besides, compared with the previous SOTA method 3D-SiamRPN \cite{3DSiamRPN}, our method performs better by a margin of $\sim$9\% and $\sim$5\% in 3D Success and 3D Precision respectively. It verifies the superiority of our method, and shows that PTT-Net could work better in challenging scenarios like sparse or occlusion scenarios, while other trackers often fail to track in these scenes.  Additionally, compared with \cite{AVOD} and \cite{FSiamese} which both use RGB+LIDAR fusion information, the Success/Precision results of PTT-Net outperform them 4.7\%/12.1\% and 30.7\%/31.2\% respectively. We think that this is because our method can capture important feature representations on tracking targets even if it is only based on raw point cloud data. More importantly, compared with other methods, our proposed method not only has a good performance, but also could run in real time with 40 FPS.

\begin{figure}[ht]
	\centering
	\setlength\abovecaptionskip{-10pt}
	\includegraphics[width=\linewidth]{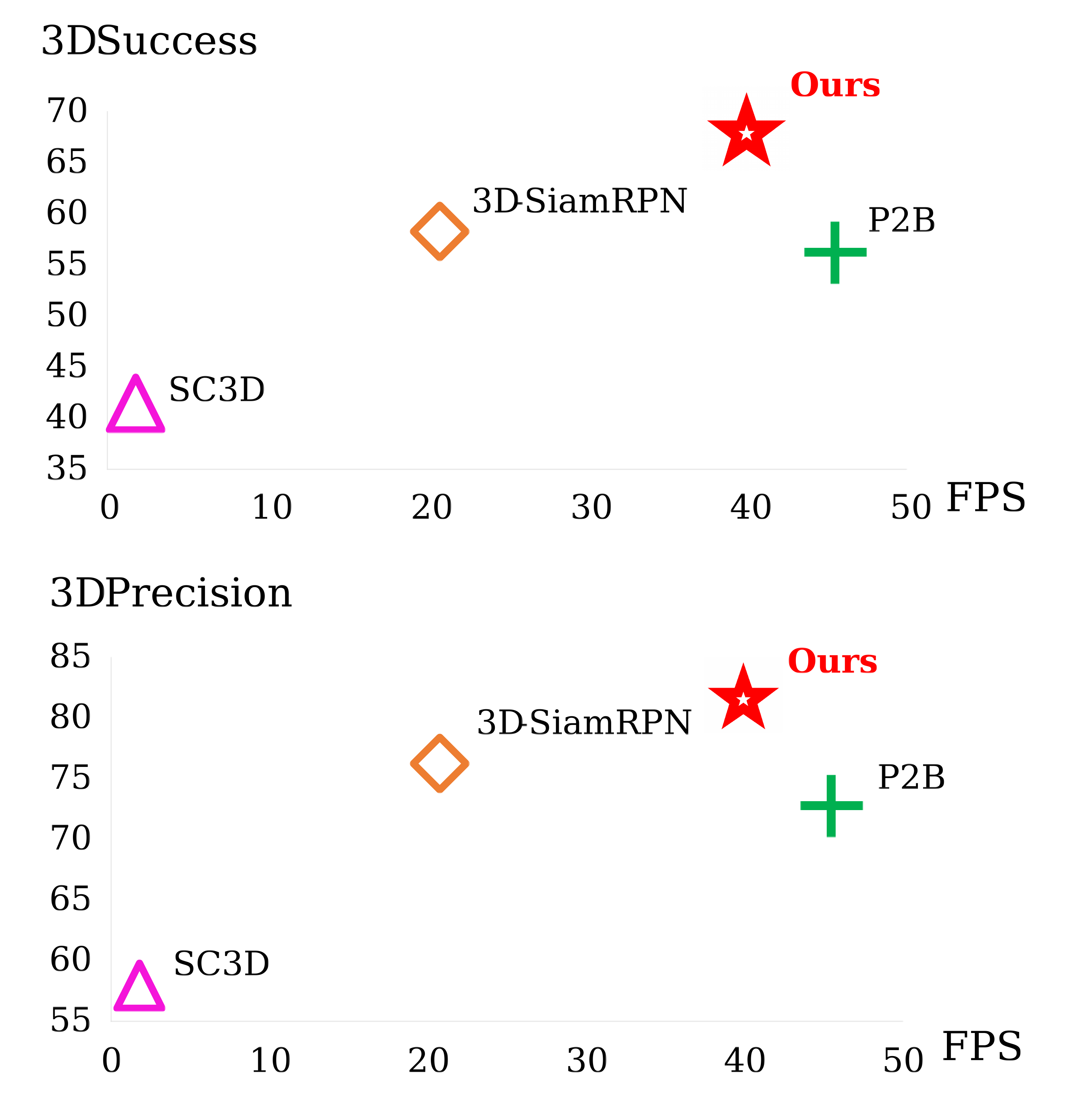}
	\caption{\textbf{Tracking Success and Precision vs. speed of the dominant trackers on Car category.} The proposed PTT-Net is superior than SC3D \cite{SC3D}, P2B \cite{P2B}, 3D SiamRPN \cite{3DSiamRPN} at 3D Success and 3D Precision, and could maintain a good inference speed.}
	\label{fig:time-performance}
\end{figure}

\begin{figure}[t!]
	\centering
	\setlength\abovecaptionskip{-10pt}
	\includegraphics[width=\linewidth]{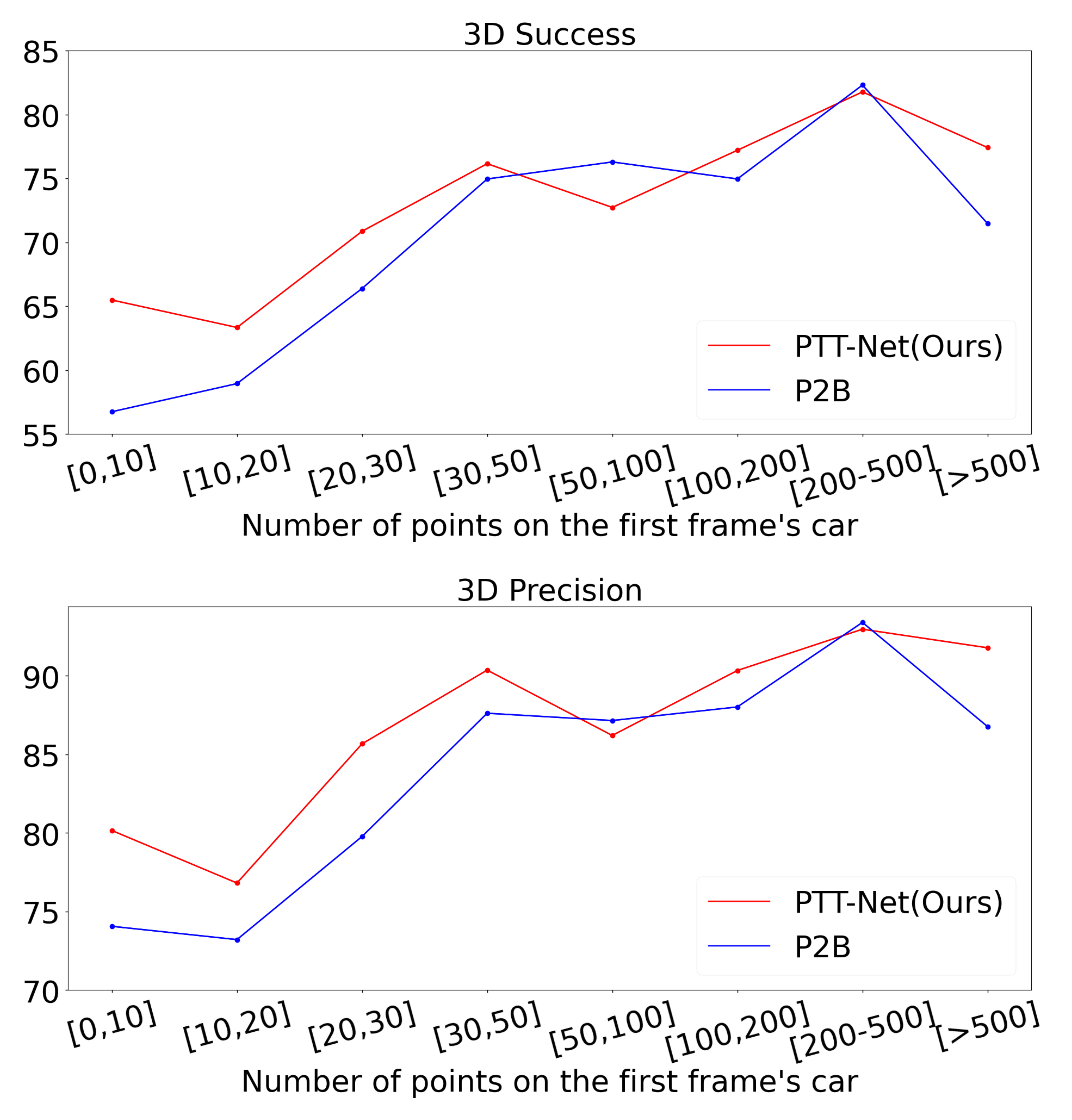}
	\caption{\textbf{The comparison of different points number intervals in the first frame between PTT and P2B}. The Success and Precision results are shown in two line charts separately. Only one case that the performance of PTT is lower than P2B when the points number interval is in [50,100]. And the average accuracy of PTT is much higher than that of P2B.}
	\label{fig:num_pts_com}
\end{figure}

\begin{figure*}[t]
	\centering
	\setlength\abovecaptionskip{-7pt}
	\includegraphics[width=\linewidth]{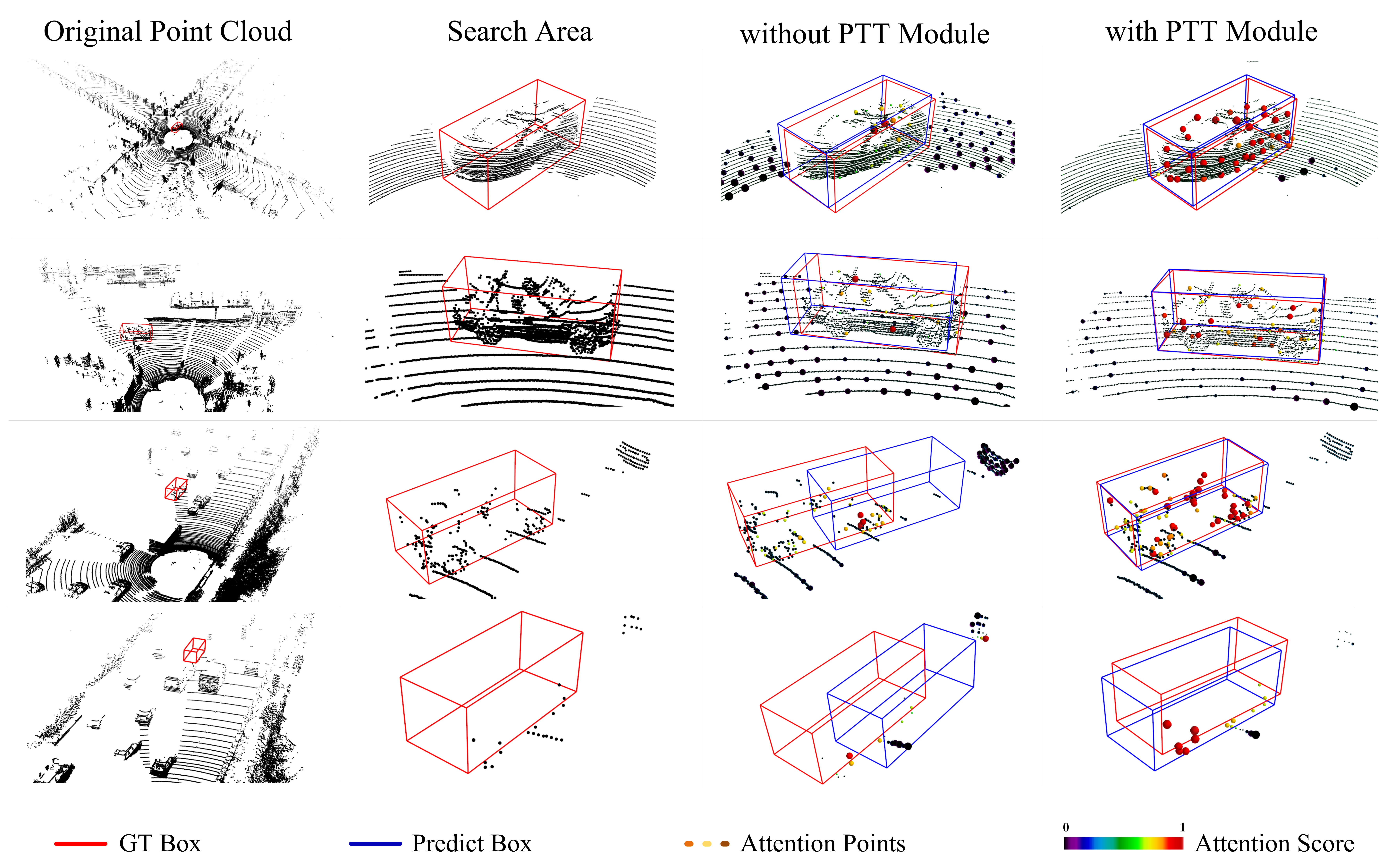}
	\caption{\textbf{Comparison of visualization results with or without PTT module.} We compared the attention points and corresponding scores in the third column and the fourth column when the network had the PTT module or not. Furthermore, the tracking difficulty is increasing as the number of initial points in the search area decreases from top to bottom.}
	\label{fig:attention-show}
\end{figure*}

\subsubsection{Comparisons on other categories}
For four categories in KITTI tracking dataset, the Car category is the rigid object. To evaluate the tracking performance in more scenes and especially containing non-rigid objects, we further compared our method with previous methods on Pedestrian, Van, and Cyclist (Tab.~\ref{tab:evaluation_KITTI}). 
Except for F-Siamese \cite{FSiamese}, which fuses image and point cloud information, all other methods adopt the same experimental settings. 

As shown in Tab.~\ref{tab:evaluation_KITTI}, the average performance of PTT-Net outperforms SC3D \cite{SC3D}, P2B \cite{P2B} and 3D-SiamRPN \cite{3DSiamRPN} by $\sim$24\%, $\sim$13\% and $\sim$9\% respectively. It is worth noting that the Success/Precision results of PTT-Net show an improvement (9.7\%/15.8\%) on non-rigid object (Pedestrian) tracking. The result also verifies that our PTT module can help the network understand and learn the important features of the target better. 
Additionally, we notice that there are performance gaps between our method and the best method F-Siamese~\cite{FSiamese} in Van and Cyclist categories. We believe there are two reasons. First, the cyclist has the least training samples (only 1529 samples for training), which may limit the performance of the transformer. And we did not do any extra data augmentation for cyclist because we would like to use a fair setting among all categories. Second, F-Siamese firstly utilizes a 2D siamese tracker in the front end only using dense image data, and generates a search area based on the results of the 2D siamese tracker, which provides prior information for subsequent 3D SOT. We believe that the data fusion in F-Siamese may result in the better performance. Additionally, we also notice that the Success results of P2B achieves 28.7\%/32.1\% in pedestrian and cyclist respectively, and our method brings +16.2\%/+5.2\% gains in the two categories. Because our method is based on P2B, we believe the results also verify the effectiveness of our method. Besides, our performance is much higher than that of F-Siamese method in the categories with more abundant data, such as vehicles and pedestrians. The Success/Precision results of PTT-Net outperform F-siamese by 30.7\%/31.2\% and 28.7\%/39.8\% in vehicles and pedestrian category respectively.

Besides, we also evaluated our PTT-Net on NuScenes dataset to further confirm the effectiveness of our method. Although KITTI dataset is commonly used by previous 3D SOT methods, its scale is too small and this may limit the performance of proposed network. Recently, BAT~\cite{bat} reports the results of the dominant methods on NuScenes dataset. Hence we follow the settings of BAT and evaluate our PTT-Net on NuScenes, and report the results in Tab~\ref{tab:evaluation_KITTI}. As shown in Tab~\ref{tab:evaluation_KITTI}, our PTT-Net outperforms BAT in all categories. This indicates that our PTT-Net could address the challenging scenes in NuScenes more effectively.

\subsubsection{Comparison of speed and performance}
We further show the comparisons with previous trackers in terms of speed, 3D Success and 3D Precision on Car category. As shown in Fig.~\ref{fig:time-performance}, our PTT-Net achieves SOTA performance on both 3D Success and 3D Precision. Meanwhile, our method has less computational burden. In other words, our method has both high accuracy and fast running speed.

\subsubsection{Comparisons in different points intervals}
Here, we choose the Car category and divide the tracking sequence according to the number of points in the first frame, which is used to initialize the template point cloud and plot the corresponding tracking performance curve (Fig.~\ref{fig:num_pts_com}). As mentioned in Sec.~\ref{sec:challenges}, the sparsity of point cloud limits the performance of 3D SOT trackers. Generally, in extremely sparse scenes (the number of points on the target is less than 50), most existing trackers often track off or fail. However, as shown in Fig.~\ref{fig:num_pts_com}, the performance of our method outperforms P2B in both Success and Precision with a large margin when points number is less than 50. This shows that our method could deal with sparse scenes better and has more robust performance.

\subsection{Qualitative Experiment}

\begin{figure*}[t!]
	\centering
	\setlength\abovecaptionskip{-10pt}
	\includegraphics[width=\linewidth]{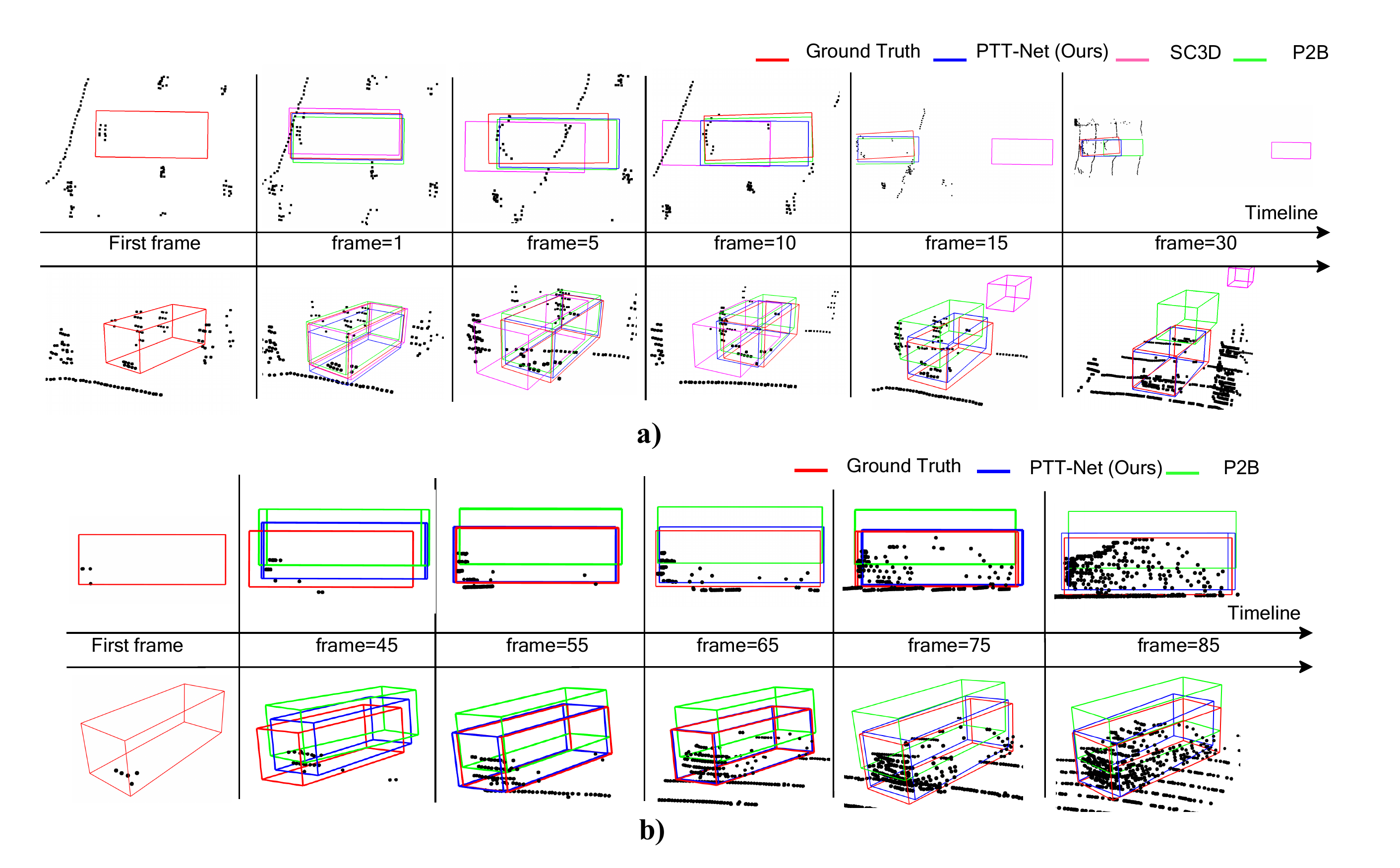}
	\caption{\textbf{Advantageous cases of PTT-Net compared with SC3D and P2B (a-b).} In (a) and (b), the number of point clouds in the first frame is less than 50. Our method can track the target accurately. However, in scenario (a), both P2B and SC3D failed to track. In scenario (b), even though P2B could track the target, it still has an inaccurate z-axis estimation for the target. Meanwhile, SC3D has failed to track. These results show the robustness of our method in sparse point cloud scenarios. Please see our experiment video for more details.}
	\label{fig:qualitative1}
\end{figure*}

\begin{figure*}[t]
\begin{center}
	\centering
	\setlength\abovecaptionskip{0pt}
	\includegraphics[width=0.95\linewidth]{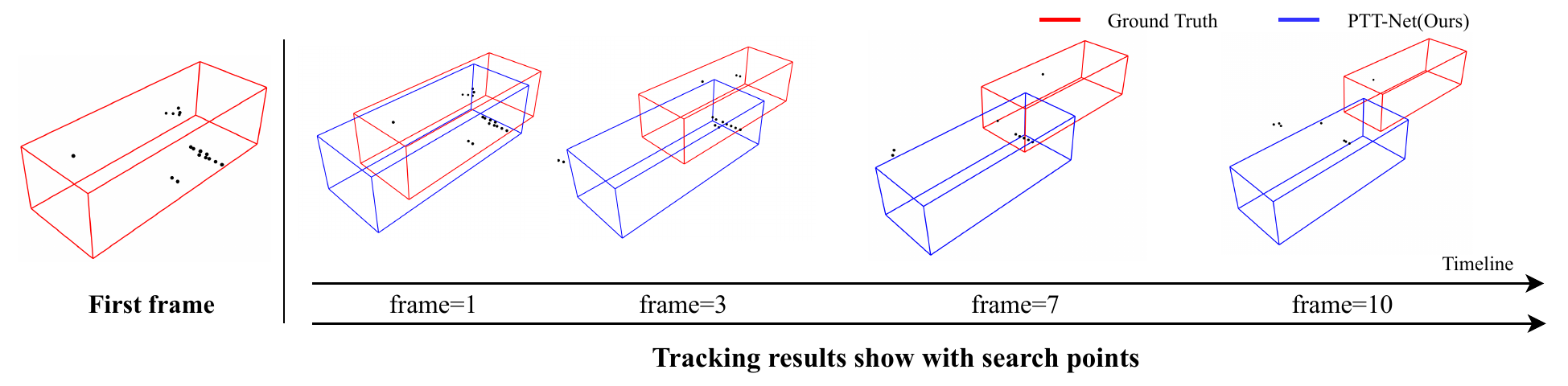}
	\caption{\textbf{Failure cases.} no points in the initial search area}
	\label{fig:failure}
\end{center}
\end{figure*}

\subsubsection{PTT Module show}
To show the effects of PTT modules in PTT-Net, we exemplified the attention points and scores when PTT-Net utilizes PTT Module or not in Fig.~\ref{fig:attention-show}. For attention points and scores, we set the voting scores in the voting stage as attention scores. The higher the score, the more attention the corresponding points have been focused on. As shown in Fig.~\ref{fig:attention-show}, from the first row to the fourth row, the point cloud of tracking target becomes more sparse, which makes the tracking difficulty higher. However, the results show that our PTT module can achieve robust tracking in all these different scenarios. And the attention scores tend to be higher in the location where the target features are rich. Meanwhile, we also observe that our PTT module helps network filter the background noise better and focus on the tracking target. Especially, from the visualization results in the fourth row of Fig.~\ref{fig:attention-show}, even if there are few points, our PTT module could still help the network focus more on the foreground points, which once again proves the power of our PTT module. 

\subsubsection{Advantageous cases}
We visualized our advantageous cases over P2B and SC3D in Fig.~\ref{fig:qualitative1}. We can observe from Fig.~\ref{fig:qualitative1} (a) that in the sparse scenarios (less than 50 points) where both SC3D and P2B tracked off course or even failed, our PTT-Net still tracks the target robustly. In Fig.~\ref{fig:qualitative1} (b), even though P2B can track the target, their position estimation still has a large deviation in z axis. This also shows the effectiveness of our method. We could not only track the target robustly in sparse scenes, but also estimate the location information of the target more accurately.

\begin{table*}
\renewcommand\arraystretch{1.3}
	\begin{center}
		\caption{Different ways for template generation. ``First \& Previous" denotes ``The first ground truth (GT) and Previous result". }
		\label{tab:different_template}
		\centering
		\begin{tabular}
		{@{\hspace{0.06pt}}
		p{0.30\columnwidth}<{\centering}
		p{0.13\columnwidth}<{\centering}
		p{0.25\columnwidth}<{\centering}
		p{0.13\columnwidth}<{\centering}
		p{0.15\columnwidth}<{\centering}
		p{0.00cm}
		p{0.13\columnwidth}<{\centering}
		p{0.25\columnwidth}<{\centering}
		p{0.13\columnwidth}<{\centering}
		p{0.15\columnwidth}<{\centering}
		@{\hspace{0.04pt}}}
			\toprule
			\textbf{Source of} & \multicolumn{4}{c}{Success } && \multicolumn{4}{c}{Precision } \\
			\cline{2-5}\cline{7-10}
			\textbf{template points} & PTT-Net & 3D-SiamRPN\cite{3DSiamRPN} & P2B\cite{P2B}  & SC3D~\cite{SC3D}  && PTT-Net & 3D-SiamRPN\cite{3DSiamRPN} & P2B\cite{P2B}  & SC3D~\cite{SC3D} \\
			\hline
			The First GT & \textbf{62.9} & 57.2       & 46.7 & 31.6 && \textbf{76.5} & 75.0       & 59.7 & 44.4 \\
			Previous result  & \textbf{64.9} & -          & 53.1 & 25.7 && \textbf{77.5} & -          & 68.9 & 35.1 \\
			First \& Preivous & \textbf{67.8} & 58.2       & 56.2 & 34.9 && \textbf{81.8} & 76.2       & 72.8 & 49.8 \\
			All previous & \textbf{59.8} & -          & 51.4 & 41.3 && \textbf{74.5} & -          & 66.8 & 57.9 \\
			\hline
		\end{tabular}
	\end{center}
\end{table*}

\begin{table*}[]
    \renewcommand\arraystretch{1.3}
    \begin{center}
    \caption{Different ways for search area generation. }
    \label{tab:different_search}
    \begin{tabular}
		{@{\hspace{0.08pt}}
		p{0.35\columnwidth}<{\centering}
		p{0.22\columnwidth}<{\centering}
		p{0.22\columnwidth}<{\centering}
		p{0.22\columnwidth}<{\centering}
		p{0.02cm}
		p{0.22\columnwidth}<{\centering}
		p{0.22\columnwidth}<{\centering}
		p{0.22\columnwidth}<{\centering}
		@{\hspace{0.04pt}}}
    \toprule
    \multirow{2}{*}{}   & \multicolumn{3}{c}{Success} &  & \multicolumn{3}{c}{Precesion} \\ \cline{2-4} \cline{6-8} 
                        & SC3D\cite{SC3D}    & P2B\cite{P2B}              & PTT-Net(Ours)               &  & SC3D\cite{SC3D}    & P2B\cite{P2B}              & PTT-Net(Ours)     \\ \hline
    Previous Result     & 41.3               & 56.2                       & \textbf{67.8}           &  & 57.9               & 72.8                       & \textbf{81.8}      \\
    Previous GT & 64.6               & \textbf{82.4 }             & 75.9                    &  & 74.5               & \textbf{90.1 }             & 88.9      \\
    Current GT  & 76.9               & \textbf{84.0}              & 76.1                    &  & 81.3               & \textbf{90.3}              & 89.1      \\ \hline
    \end{tabular}
    \end{center}
\end{table*}
\subsubsection{Failure cases}
The Fig.~\ref{fig:qualitative1} shows that PTT-Net can work well in most of scenes compared to SC3D and P2B. However, the sparsity of points still influences the performance of PTT-Net. Our method tends to fail when the points are extremely sparse. To show the failure case of our method in more detail, we visualized the failed tracking result when the points are less than 20. As shown in Fig.~\ref{fig:failure}, our PTT-Net could not learn effective object features since there are almost no points in the initial search area. In addition, due to the sparse point cloud, the feature ambiguity also leads to the inaccurate estimation of the bounding box, which causes the propagation of errors and failure case finally.

\subsection{Ablation Study}
\label{sec:ablation-study}

Here, we ablate the network architecture on KITTI dataset. First, we discuss different ways for seeds sampling, template generation, and search area generation. Then we ablate the different embedded position of our PTT module. Finally, different parameters selection of PTT module are also discussed.

\subsubsection{Ways for seeds sampling}
\label{sec:sampling}
The first ablation study presented is designed to support our claim that FPS could benefit the classification task. As mentioned in Sec.~\ref{sec:sample}, different sampling methods will lead to different degrees of target information loss. Here, we compared the effects of three different down-sampling methods: Random Sampling (RS), Farthest Point Sampling (FPS), and Feat-FPS proposed by \cite{3DSSD} (Tab.~\ref{tab:diff_sample_results}) on the performance of our method. We also showed the visualizations of three sampling methods (Fig.~\ref{fig:sample_method}). We found that RS had the worst performance, and FPS could obtain the best performance, which was $\sim$7\% higher than RS. Besides, Feat-FPS also had good tracking performance, which was only $\sim$1\% lower than the FPS. We attribute this result to the fact that FPS can obtain seeds which belong to the foreground and background points uniformly. Meanwhile, FPS could keep the distribution probability of the original input point cloud to the greatest extent while reducing the dimension of the input, which will be beneficial to the classification and regression tasks of the tracking network.

\begin{table}[t!]
\renewcommand\arraystretch{1.3}
    \centering
    \begin{center}
    \caption{Performance of different sampling methods.}
    \label{tab:diff_sample_results}
    \setlength\tabcolsep{13.0pt}{
    \begin{tabular}{cccc}
    \toprule
    {} & Random Sample & Feat-Fps & Fps \\ \hline
    3D Success & 60.4 & 66.1 & \textbf{67.8}    \\
    3D Precision & 73.7 & 80.0 & \textbf{81.8} \\
    \hline
    \end{tabular}}
    \end{center}
\end{table}

\begin{table}[t!]
\renewcommand\arraystretch{1.3}
    \begin{center}
    \caption{Different embedded positions of PTT module.}
    \label{tab:Ablation-ptt-pos}
	\centering
	\setlength\tabcolsep{15.0pt}{
	\begin{tabular}{ccc} 
	    \toprule
        \textbf{~Ablation} & 3D Success & 3D Precision \\
        \midrule
        baseline\cite{P2B}  & 56.2 & 72.8 \\
        Only PTT in Vote & 62.1& 76.9 \\
        Only PTT in Prop & 65.7 & 78.9 \\
        PTT in all (PTT-Net) &\pmb{67.8} & \pmb{81.8} \\ \hline
    \end{tabular}}
    \end{center}
\end{table}

\subsubsection{Ways for template generation}
Our method is consistent with P2B when generating template point clouds. Therefore, we explored the different way of template point cloud generation, including the first ground truth, previous result, the fusion of the first ground truth and previous result, and all previous results. Specially, the fusion between first frame ground truth and previous results means the fusion of two point clouds within the two 3D bounding boxes respectively. First, we extract the points in the box from the point cloud according to the first frame ground-truth box and the predicted box in the previous frame. Second, according to the angle of the box and the coordinates of the center point, the two frames of point clouds are normalized to the same coordinate system by rotating and translating respectively. Finally, the updated template point clouds could be obtained by concatenating two point clouds directly. We reported the results in Tab.~\ref{tab:different_template}. We found that our PTT-Net outperformed 3D-SiamRPN, P2B and SC3D in all settings. Besides, the proposed method has the best performance by fusing the first ground truth and the previous result. We believe that this is because our method has more robust tracking results in sparse scenes. Therefore, if we fuse the first ground truth and the previous result to update the template points, it could further enhance the target information and improve the algorithm performance. 

\subsubsection{Ways for search area generation}
The generation strategy of search area in object tracking task directly determines the feature scale and quality that the network can learn. In previous work SC3D \cite{SC3D}, P2B \cite{P2B}, there are performance comparisons in different search area situations. To further explore the performance of our method, we also conducted experiments on the search area generation strategies, and compared them with SC3D \cite{SC3D} and P2B \cite{P2B}. The experimental results are shown in Tab.~\ref{tab:different_search}. Specifically, we compared three different search area generation methods: 1) centered on previous result; 2) centered on previous ground truth; 3) centered on current ground truth. 
The results show that the performance of the three methods has been greatly improved with search area generated by ground-truth. The reason is that each frame uses the ground-truth result, which can effectively avoid the accumulation of errors caused over time. However, it is worth noting that our method PTT is slightly lower than P2B after using the ground-truth, but the performance is still at a similar level.

\subsubsection{Positions for PTT module embedding}
To verify our design in Sec.~\ref{sec:ptt-net} of positions where PTT modules are embedded, we tried different schemes (Tab.~\ref{tab:Ablation-ptt-pos}). The results show that embedding PTT module in both stages of \cite{P2B} can obtain the best improvement. 
Comparing (a) with (b) in Fig.~\ref{fig:transformer_show}, PTT-Net has better point cloud classification results which focus on foreground points. Comparing (c) with (d), PTT-Net could still track the target pedestrian robustly when more proposal centers are generated from another pedestrian. Besides, as shown in Fig.~\ref{fig:attention-show}, PTT-Net can focus on foreground points with the help of PTT module. This result effectively shows that the transformer can learn more target-wise information.

\subsubsection{Parameters selection for PTT module}
Here we discuss the details of our PTT module, including the number of heads and the number of attention layers, as shown in Tab.~\ref{tab:Ablation-hyperparams}. For the number of heads, we observe that $head = 1$ and $layer = 1$ achieves the best performance, and stacking more heads or layers could not bring in performance improvement but more parameters and lower speed. We believe that since our PTT module is directly applied on the fusion feature, which already has the similarity representations, more heads or layers in PTT may make the feature focus on other unimportant features, thus distracting the already fused similarity features. Therefore, we set both the heads and layers to 1.

\begin{table}[]
\renewcommand\arraystretch{1.3}
    \begin{center}
    \caption{Ablation study of hyper-parameters in PTT module.}
    \label{tab:Ablation-hyperparams}
	\centering
	\setlength\tabcolsep{15.0pt}{
	\begin{tabular}{cccc} 
	    \toprule
        \textbf{} & {} & 3D Success & 3D Precision \\
        \midrule
        \multirow{4}{*}{Head Number}
        &1 &\textbf{67.8} & \textbf{81.8}\\
        &2 &67.2 & 80.9\\
        &4 &65.6 & 79.1\\
        &8 &65.5 & 78.1\\\hline
        \multirow{4}{*}{Layer Number}
        &1 &\textbf{67.8} & \textbf{81.8}\\
        &2 &64.7 & 78.5\\
        &4 &64.0 & 76.8\\
        &8 &62.6 & 76.5\\\toprule
    \end{tabular}}
    \end{center}
\end{table}

\subsection{Timing Breakdow}
We calculated the average running time of all test frames for car to measure PTT-Net's speed. PTT-Net achieved 40 FPS on a single NVIDIA 1080Ti GPU, including 8.3 ms for preparing point cloud, 16.2 ms for model forward propagation, and 0.5 ms for post-processing. The running time of SC3D \cite{SC3D}, P2B \cite{P2B} and 3D-SiamRPN \cite{3DSiamRPN} on the same platform are 1.8FPS, 45.5FPS and 20.8FPS, respectively.

\section{Conclusions}
In this work, we explored the application of transformer network in 3D SOT task and proposed PTT module. The PTT module aims at weighing point cloud features to focus on the important features of objects. We also embedded the PTT module into the open-source state-of-the-art method \cite{P2B} to build a PTT-Net 3D tracker. Experiments show that PTT-Net outperforms previous state-of-the-art methods with remarkable margins. We hope that our work will inspire further investigation of the application of transformers to 3D object tracking.

\section*{ACKNOWLEDGMENT}
This work was supported by National Natural Science Foundation of China (62073066, U20A20197), Science and Technology on Near-Surface Detection Laboratory (6142414200208), the Fundamental Research Funds for the Central Universities (N182608003), Major Special Science and Technology Project of Liaoning Province (No.2019JH1/10100026), and Aeronautical Science Foundation of China (No.201941050001).
\ifCLASSOPTIONcaptionsoff
  \newpage
\fi


\bibliographystyle{IEEEtran}
\bibliography{tmm_final}

\begin{IEEEbiography}[{\includegraphics[width=1in,height=1.25in,clip,keepaspectratio]{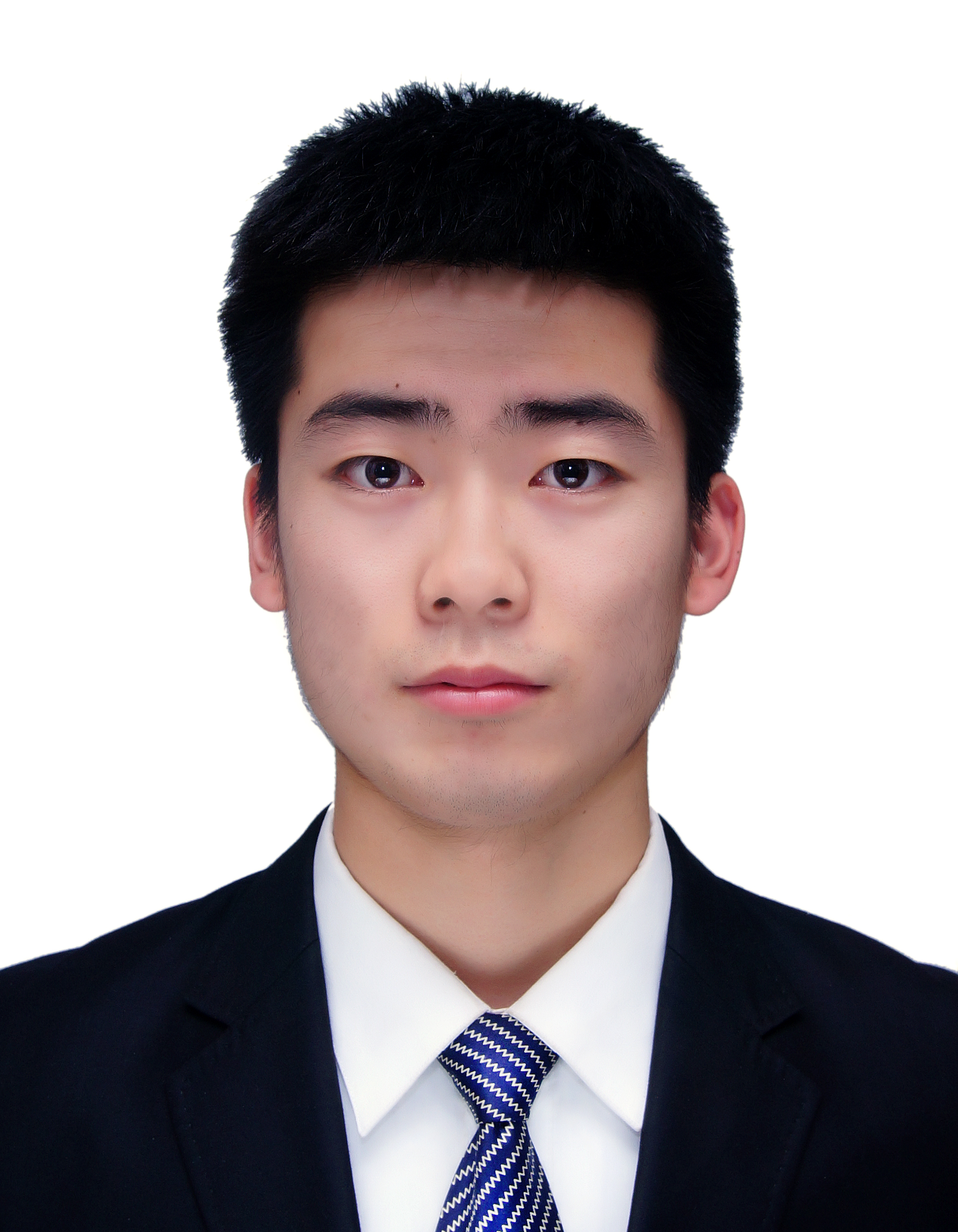}}]{Jiayao Shan}
received the B.S. degree from the Dalian Jiaotong University of China. He is currently pursuing the M.S. degree from the Northeastern University. His research interests include deep learning, lidar perception and 3D object detection and tracking.
\end{IEEEbiography}

\begin{IEEEbiography}[{\includegraphics[width=1in,height=1.25in,clip,keepaspectratio]{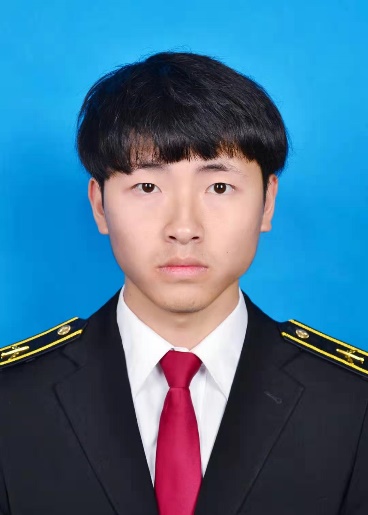}}]{Sifan Zhou}
received the B.S. degree from the Civil Aviation University of China. He is currently pursuing the M.S. degree from the Northeastern University. His research interests include deep learning, visual perception and 3D object tracking.
\end{IEEEbiography}

\begin{IEEEbiography}[{\includegraphics[width=1in,height=1.25in,clip,keepaspectratio]{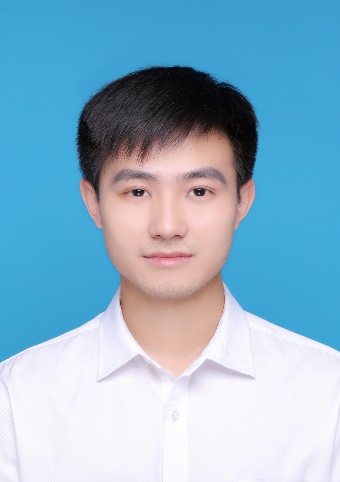}}]{Yubo Cui}
received the B.S. degree in Automation and the M.S. degree in Robot Science and Engineering from Northeastern University, China in 2017 and 2020 respectively. His research interests include deep learning, 3D object tracking, and 3D detection.
\end{IEEEbiography}

\newpage

\begin{IEEEbiography}[{\includegraphics[width=1in,height=1.25in,clip,keepaspectratio]{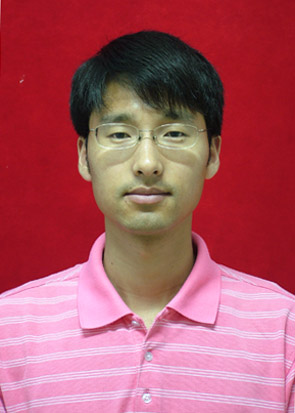}}]{Zheng Fang}
received the B.S. degree in Automation and Ph.D. degree in Pattern Recognition and Intelligent Systems from Northeastern University, China in 2002 and 2006 respectively. He was a postdoctoral research fellow at Carnegie Mellon University from 2013 to 2015. He is now a full professor in Faculty of Robot Science and Engineering at Northeastern University, China. His research interests include visual/ Laser SLAM, perception and autonomous navigation of various mobile robots.
\end{IEEEbiography}

\end{document}